\documentclass[12pt]{article}
\usepackage{amsmath}
\usepackage{graphicx}
\usepackage{enumerate}
\usepackage{natbib}
\usepackage{url} % not crucial - just used below for the URL
\usepackage{color}
\usepackage{tabularray}
\usepackage{tabularx}
\usepackage{xcolor}
\usepackage{adjustbox}
\usepackage{tikz}
\usepackage{booktabs}
\usepackage{amssymb}
\usepackage{bbding}
\usepackage{pifont}
\usepackage{wasysym}
\usepackage{utfsym}
\usepackage{fontawesome}
\usepackage{enumitem}
\usepackage{listings}
\usepackage{amsmath}
\usepackage{tcolorbox}
\usepackage{amsmath}
\usepackage{graphicx,psfrag,epsf}
\usepackage{enumerate}
\usepackage{natbib}
\usepackage{url} % not crucial - just used below for the URL
\usepackage{algorithm}
\usepackage{algpseudocode}
\usepackage{multirow}
\usepackage{authblk}
\usepackage{caption}
\usepackage{tikz}
\usepackage{threeparttable}

\usepackage{url}            % simple URL typesetting
\usepackage{booktabs}       % professional-quality tables
\usepackage{amsfonts}       % blackboard math symbols
\usepackage{mathrsfs}
\usepackage{nicefrac}       % compact symbols for 1/2, etc.
\usepackage{microtype}      % microtypography
\usepackage{xcolor}         % colors
\usepackage{graphicx}
\usepackage{epstopdf}
\usepackage{float}
\usepackage{subfigure}
\usepackage{amssymb}
\usepackage{bbding}
\usepackage{setspace}

\newcommand{\blind}{1}

\begin{document}

\def\spacingset#1{\renewcommand{\baselinestretch}%
{#1}\small\normalsize} \spacingset{1}

%%%%%%%%%%%%%%%%%%%%%%%%%%%%%%%%%%%%%%%%%%%%%%%%%%%%%%%%%%%%%%%%%%%%%%%%%%%%%%

\if1\blind
{
  \title{\bf A Survey on Large Language Model-based Agents for Statistics and Data Science}
  \iffalse
  \author{Author 1\thanks{
    The authors gratefully acknowledge \textit{please remember to list all relevant funding sources in the unblinded version}}\hspace{.2cm}\\
    Department of YYY, University of XXX\\
    and \\
    Author 2 \\
    Department of ZZZ, University of WWW}
  \maketitle
}
\fi

\author{Maojun $\rm{Sun}^{a}$,  Ruijian $\rm{Han}^{a}$,
	Binyan $\rm{Jiang}^{a}$,
Houduo  $\rm{Qi}^{a,b}$,\\
Defeng $\rm{Sun}^{b}$,
	Yancheng $\rm{Yuan}^{b*}$
	and
	Jian $\rm{Huang}^{a,b}$\thanks{Corresponding authors.}
	
	\vspace{0.2cm}
{\footnotesize { {$\it^{a}$Department of Data Science and Artificial Intelligence, The Hong Kong Polytechnic University}}}\\
	{\footnotesize {{$\it^{b}$Department of Applied Mathematics, The Hong Kong Polytechnic University}}}
}
\date{}
  \maketitle}

 \fi

\if0\blind
{
  \bigskip
  \bigskip
  \bigskip
  \begin{center}
    {\LARGE\bf A Survey on Large Language Model-based Agents for Statistics and Data Science}
\end{center}
  \medskip
} \fi

\bigskip

\begin{abstract}
In recent years, data science agents powered by Large Language Models (LLMs), known as ``data agents," have shown significant potential to transform the traditional data analysis paradigm. This survey provides an overview of the evolution, capabilities, and applications of LLM-based data agents, highlighting their role in simplifying complex data tasks and lowering the entry barrier for users without related expertise. We explore current trends in the design of LLM-based frameworks, detailing essential features such as planning, reasoning, reflection, multi-agent collaboration, user interface, knowledge integration, and system design, which enable agents to address data-centric problems with minimal human intervention. Furthermore, we analyze several case studies to demonstrate the practical applications of various data agents in real-world scenarios. Finally, we identify key challenges and propose future research directions to advance the development of data agents into intelligent statistical analysis software.
\end{abstract}

\noindent%
{\it Keywords: data agents; generative AI; data analysis; natural language interaction; statistical software.}  %3 to 6 keywords, that do not appear in the title
\vfill

\newpage
\spacingset{1.9} % DON'T change the spacing!
\section{Introduction}
\label{sec:intro}
As nearly every aspect of society becomes digitized, data analysis has emerged as an indispensable tool across various industries \citep{inala2024data}. For instance, financial institutions leverage data analysis to make informed decisions about stock trends \citep{provost2013data,mckinsey2011big}, hospitals utilize it to monitor patients' health conditions \citep{waller2016data}, and companies employ it to develop strategic plans \citep{chen2012business}. Despite its widespread utility, data analysis is often perceived as a challenging field with a significant ``entry barrier" \citep{cao2017data,jordan2015machine}, typically requiring knowledge in areas such as statistics, data science, and computer science \citep{kitchin2014big}. Since the release of SPSS \citep{SPSS} in 1968, followed by SAS \citep{SAS}, Matlab \citep{Matlab}, Excel \citep{Excel}, Python \citep{Python}, R \citep{R}, PowerBI \citep{PowerBI}, and other specialized data analysis tools and programming languages, these advancements have significantly aided professionals in conducting statistical experiments and data analysis. Moreover, they have made data analysis more accessible to a broader range of practitioners \citep{witten2016data}.

The general data analysis process typically involves several key steps. Initially, data is collected from studies or extracted from databases and imported into tools such as Excel. Next, software like Excel or programming languages such as Python and R are employed to clean and analyze the data, aiming to extract valuable insights. Subsequently, data visualization is performed to make these insights more accessible and understandable. For more complex tasks, such as statistical inference and predictive analysis, statistical and machine learning models are often necessary. This involves data processing, feature engineering, modeling, evaluation, and more. Upon completing the analysis, a final report is usually drafted to summarize the findings and insights. However, for individuals without expertise in statistics, data science, and programming, data analysis remains a high-barrier task.

The barriers to data analysis primarily exist in the following areas:

\noindent

\begin{itemize}
\item \text{Lack of systematic statistical training}:
    Individuals without a background in statistics may find it challenging to understand which types of analysis are feasible, even when data is presented to them. As data and models become increasingly complex, gaining a solid understanding of current statistical techniques typically requires at least a Master's level of statistical training.

\item \text{Software limitation}: Simple data analysis tools like Excel are inadequate for complex scenarios, such as predictive analysis or analyzing data from enterprise databases. Conversely, advanced programming languages for data analysis, such as Python and R, require prior programming knowledge, which can be a barrier for many users.

\item \text{Challenges in domain-specific problems}:  In specialized fields like protein or genetic data analysis, general data scientists may find it difficult to perform effective analysis due to a lack of domain-specific knowledge.

\item \text{Difficulty in integrating domain knowledge}: Corresponding to the last point, domain experts often lack the data science and programming skills needed to quickly incorporate their expertise into data analysis tools. For example, PSAAM \citep{PSAAM} is software designed for the curation and analysis of metabolic models, yet a biologist researching metabolism might find it challenging to integrate this analytical method into common data analysis tools like Excel or R.
\end{itemize}

With the rise of generative AI, new opportunities have emerged in statistics and data science. LLM-based data agents are gradually addressing existing challenges while introducing a new paradigm for approaching data analysis tasks.

An ``AI agent" (or LLM agent) refers to an autonomous or semi-autonomous software system powered by AI models such as LLMs. These agents can interpret natural language instructions, plan and execute tasks, and interact with users or other systems to complete complex workflows \citep{cheng2024exploring}.

Specifically, we define an LLM-based data agent as an autonomous or semi-autonomous software system powered by LLMs, capable of understanding natural language instructions, planning and executing data-centric tasks, and interacting with users or external tools to accomplish complex objectives—from exploratory data analysis to machine learning model development. In this paper, the terms ``LLM-based data science agent," ``LLM-based data agent," and ``data science agent"  are collectively referred to as “data agent” for simplicity.

This survey explores recent advancements in data agents and highlights data analysis performed by various agents through a series of case studies. In Section \ref{ai1}, we briefly discuss the opportunities introduced by recent developments in generative AI. Section \ref{sec:data agent} reviews and categorizes recent work on data science agents. We then present several case studies in Section \ref{sec:case_study}. Section \ref{sec:future} examines the challenges and future directions in this field, followed by our discussion in Section \ref{sec:diss}. Finally, we present our conclusions in Section \ref{sec:con}.

\section{Opportunities Brought by Generative AI}\label{ai1}
The rise and potential of generative AI, particularly Large Language Models (LLMs) or vision language models (VLMs) in the field of data science and analysis have gained increasing recognition in recent years. In addition to understand text, LLMs are also trained to understand tabular data, allowing them to effectively extract insights, identify patterns, and draw meaningful conclusions from tables \citep{10.1145/3626772.3661384}. Consequently, LLMs have emerged as powerful tools capable of significantly enhancing and transforming a variety of data-driven applications and workflows \citep{nejjar2023llms, tu2023datascienceeducationlarge, cheng2023gpt}. Recent research has focused on designing LLM-based data science agents (data agents) to automatically address data science tasks through natural language, as demonstrated by tools like ChatGPT-Advanced Data Analysis (ChatGPT-ADA) \citep{openai2024gpt4}, LAMBDA \citep{sun2024lambda} and Colab Data Science Agent \citep{googleblog2024datascienceagent}.

 The emergence of data agents offers a potential solution to the previously mentioned challenges, as they lower the entry barrier for users who lack programming or statistical knowledge. By providing an intuitive interface that harnesses the capabilities of LLMs, users can request analyses using natural language, and the data agents can interpret these instructions, access relevant data, and autonomously apply appropriate analytical techniques. For example, a user might request, “Calculate the sales growth in different regions from 2021 to 2028, generate a bar chart to visualize the results, and provide key insights.” With this simplified instruction, data agents can automatically extract, analyze, visualize, and report data, reducing the requirement for technical expertise and fostering a more efficient workflow. This significantly lowers the entry barriers for individuals unfamiliar with traditional data analysis tools and methods.

Furthermore, by embedding specialized knowledge into LLMs, data agents can potentially overcome challenges faced by data scientists in fields like genomics, where domain expertise is crucial \citep{cao2017data}. Simultaneously, domain experts who may lack data science or programming skills can rely on data agents to seamlessly integrate their expertise into data analysis workflows. This ability to bridge the gap between domain expertise and data science has the potential to advance interdisciplinary research and decision-making in complex scenarios.

\begin{figure}[h]
	\centering
	\includegraphics[width=\textwidth]{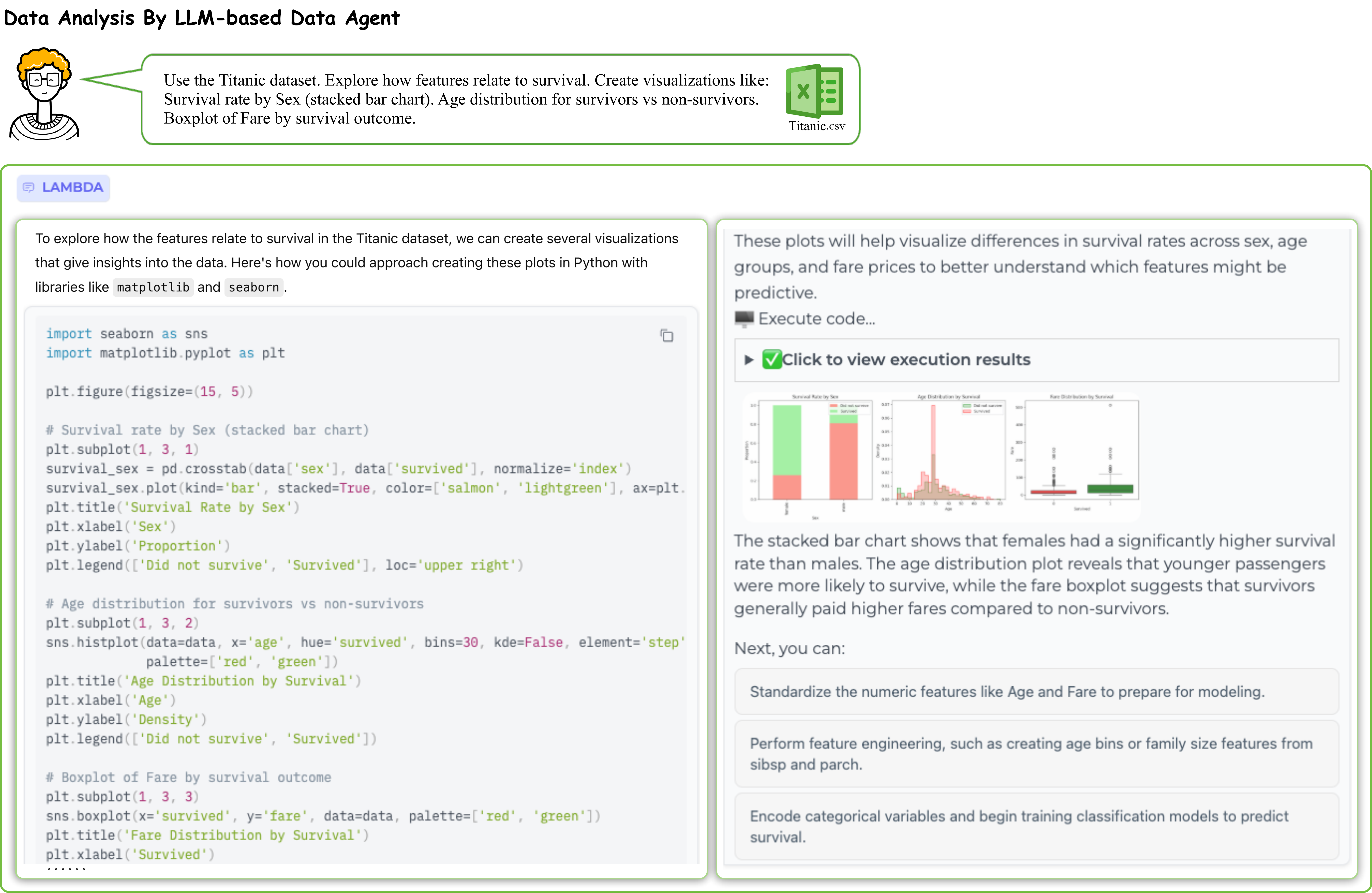}
	\caption{\label{traditional_vs_data_agent}New paradigm of data analysis brought by generative AI.}
\end{figure}

\section{LLM-based Data Science Agents}\label{sec:data agent}
\subsection{Overview}
\begin{figure}[h]
	\centering
	\includegraphics[width=\textwidth]{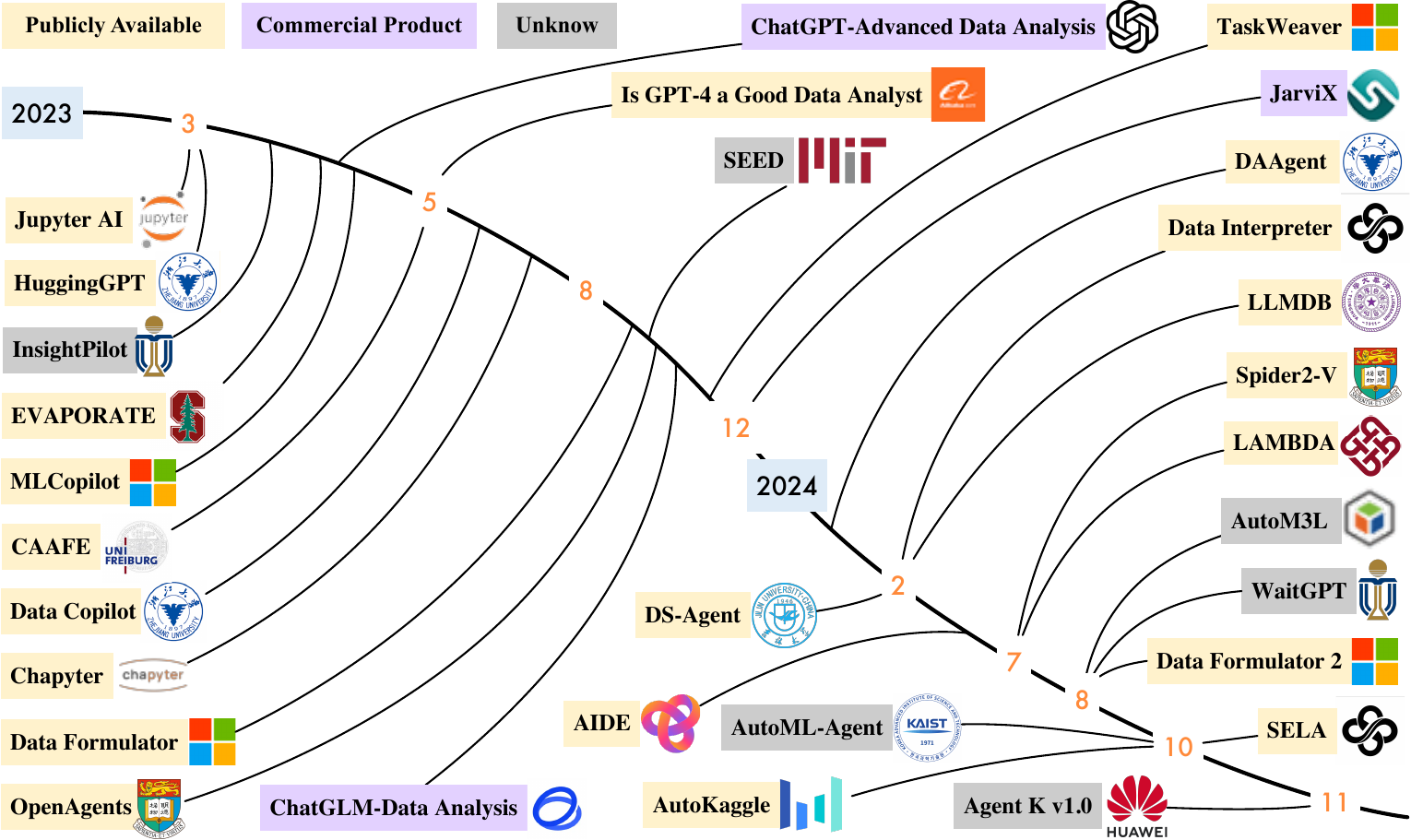}
	\caption{\label{fig:data-agent} Timeline of selected related works from 2023.}
\end{figure}

\begin{figure}[h]
	\centering
	\includegraphics[width=\textwidth]{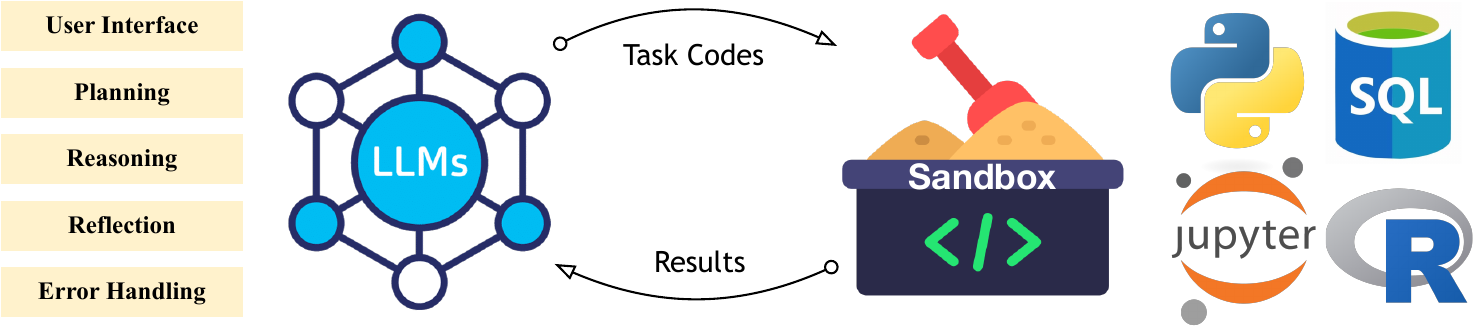}
	\caption{\label{fig:framework} An architecture of an LLM-based data agent. The diagram illustrates the interaction between LLMs and a sandbox environment. On the left, key components of LLMs are highlighted, including User Interface, Planning, Reasoning, Reflection, and Error Handling. The sandbox, positioned centrally, serves as a controlled environment for executing task codes and generating results. On the right, various tools and software that can be pre-installed in the sandbox, such as Python, SQL, Jupyter, and R, indicate the diverse ecosystems where LLM-powered agents can operate.}
\end{figure}

LLM-based data agents leverage the powerful natural language understanding and generation capabilities of LLMs to autonomously tackle complex data analysis tasks. Figure \ref{fig:framework} illustrates a commonly used framework for these agents.

In this framework, the LLM serves as the core of the entire system, driving its performance and reliability. As such, the capabilities of the LLM are critical to the system’s effectiveness, with advanced models like GPT-4 often being used. Data analysis typically involves multiple steps, especially when addressing complex tasks. Techniques such as Planning, Reasoning, and Reflection help ensure that the LLM processes these tasks with greater logical coherence and makes optimal use of its knowledge.

In the architecture, the LLM generates the code for a given data analysis task, executes it, and retrieves the corresponding results. This requires an execution environment, represented by the Sandbox, which safely isolates the code execution process. The Sandbox allows users to run programs and access files without risking the underlying system or platform. It includes pre-installed programming environments and software, such as Python, R, Jupyter, and SQL Server.

A user-friendly interface is also essential to improving usability. An intuitive interface not only attracts users but also enables them to quickly engage with and utilize the system effectively.

\subsection{Evolution of Data Science Agent}

Research on data agents began gaining momentum in 2023. \cite{chandel2022training} trained and evaluated a model within a Jupyter Notebook to predict code based on given commands and results. Soon after, it was discovered that LLMs, such as GPT, could generate accurate code for basic data analysis. With the rise of the LLM-based agent, researchers began designing special data agents for automating data science and analysis tasks by human language. Figure \ref{fig:data-agent} shows some selected works from 2023, while Table \ref{tab:data_agents} illustrates some key characteristics.

\begin{table}[h]
    \centering
    \caption{Characteristics of selected data agents. Methods can be categorized into Conversational and End-to-End approaches. Conversational methods support interactive dialogue with iterative user feedback, whereas End-to-End approaches rely on a single prompt, with the agent autonomously planning and solving the problem. The user interface can be categorized into IDE-based, Systems, CLI, and OS-based. The term ``Human-in-the-Loop" indicates that humans can intervene in the data agent's workflow, such as modifying code in situations where automatic processes are inadequate. ``Self-Correcting" refers to the agent's ability to automatically identify and correct errors within the workflow through reflection. Finally, ``Expandable" denotes the data agent's capacity to incorporate customized tools or knowledge. ``-" indicates that the attribute is either not mentioned in the paper or could not be observed from the provided resources.}\label{tab:data_agents}
    \begin{adjustbox}{width=\textwidth , keepaspectratio}
    \begin{tblr}{
      colspec = {c c c c c c c},
      hline{1,25} = {-}{0.08em},
      hline{2} = {-}{}
    }
    Data Agents           & Methods         & User Interface     & Planning     & Human in the Loop    & Self-correcting    & Expandable    \\
    ChatGPT-ADA \citep{openai2024gpt4}  & Conversational         & System    & Linear      & \textcolor{red}{\ding{55}}     & \textcolor{teal}{\ding{52}}  & \textcolor{red}{\ding{55}}    \\
    Data Copilot \citep{zhang2023data}  & End-to-end          & System    & Linear        & \textcolor{red}{\ding{55}}      & \textcolor{teal}{\ding{52}}   & \textcolor{red}{\ding{55}}  \\
    Jupyter AI \citep{jupyterlab}   & Conversational           & IDE-based   & Basic IO     & \textcolor{teal}{\ding{52}}    & \textcolor{red}{\ding{55}}    & \textcolor{red}{\ding{55}}  \\
    %InsightPilot \citep{ma2023insightpilot}  & -                & -                   & -  \\
    MLCopilot \citep{zhang2023mlcopilot}  & Conversational        & IDE-based      & Basic IO     & \textcolor{teal}{\ding{52}}    & \textcolor{red}{\ding{55}}    & \textcolor{red}{\ding{55}}  \\
    % HuggingGPT \citep{shen2024hugginggpt}  & End-to-end         &  \\
    % EVAPORARE \citep{arora2023language} \\
    % CAFFE \cite{hollmann2024large} \\
    Chapyter \citep{chapyter}   & Conversational            & IDE-based     & Basic IO      & \textcolor{teal}{\ding{52}}    & \textcolor{red}{\ding{55}}    & \textcolor{red}{\ding{55}}    \\
    Openagents \citep{xie2023openagents}   & Conversational      & System   & Linear    & \textcolor{red}{\ding{55}}      & \textcolor{red}{\ding{55}}  & \textcolor{teal}{\ding{52}}  \\
    JarviX \citep{chen2310seed} & End-to-end  & -   & -  & -  & - & - \\
    %DAAgent \citep{hu2024infiagent} & End-to-end   & CLI  & Linear  & \textcolor{red}{\ding{55}}   & -   & \textcolor{red}{\ding{55}}    \\
    DS-Agent \citep{guo2024ds} & End-to-end  & CLI  & Linear  & \textcolor{red}{\ding{55}}   & \textcolor{teal}{\ding{52}}  & -  \\
    %LLMDB \citep{zhou2024llm} & Conversational  & System  & Linear & \textcolor{red}{\ding{55}} & \textcolor{teal}{\ding{52}}  & -  \\
    Spider2-V \citep{Spider2-V} & End-to-end     & OS-Based    & -    & \textcolor{red}{\ding{55}}     & \textcolor{teal}{\ding{52}}   & - \\
    ChatGLM-DA \citep{glm2024chatglm} & Conversational  & System   & Linear    & \textcolor{red}{\ding{55}}      & \textcolor{teal}{\ding{52}}  & \textcolor{red}{\ding{55}} \\
    TaskWeaver \citep{qiao2023taskweaver} & End-to-end        & CLI \& System   & Linear  & \textcolor{red}{\ding{55}}      & \textcolor{teal}{\ding{52}}  & \textcolor{teal}{\ding{52}}  \\
    Data Interpreter \citep{hong2024data} & End-to-end       & CLI     & Hierarchical            & \textcolor{teal}{\ding{52}}      & \textcolor{teal}{\ding{52}}  & \textcolor{teal}{\ding{52}}   \\
    LAMBDA \citep{sun2024lambda}   & Conversational      & System   & Basic IO   & \textcolor{teal}{\ding{52}}    & \textcolor{teal}{\ding{52}}  & \textcolor{teal}{\ding{52}}   \\
    Data Formulator 2 \citep{wang2024data} & Conversational    & System   & Basic IO  & \textcolor{red}{\ding{55}}    & \textcolor{teal}{\ding{52}}  & -  \\
    AutoM3L \citep{luo2024autom3l} & End-to-end  & -   & -  & \textcolor{red}{\ding{55}}  & -  & \textcolor{teal}{\ding{52}}  \\
    % WaitGPT \citep{xie2024waitgpt} & End-to-end  & -   & -  & \textcolor{red}{\ding{55}}  & - & - & - \\
    SELA \citep{chi2024sela} & End-to-end  & CLI   & Hierarchical  & \textcolor{red}{\ding{55}}  & \textcolor{teal}{\ding{52}}   & -  \\
    % DS-Agent \citep{guo2024ds} & End-to-end  & Command   & Linear  & -  & \textcolor{teal}{\ding{52}} & \textcolor{teal}{\ding{52}} \\
    AIDE \citep{jiang2024aide} & End-to-end  & CLI   & Hierarchical  & \textcolor{red}{\ding{55}}  & \textcolor{teal}{\ding{52}} & - \\
    AutoKagle \citep{li2024autokaggle} & End-to-end  & CLI   & Linear  & \textcolor{teal}{\ding{52}}  & \textcolor{teal}{\ding{52}}  & \textcolor{teal}{\ding{52}} \\
    AutoML-Agent \citep{trirat2024automl} & End-to-end  & -   & Linear  & -  & \textcolor{teal}{\ding{52}}  & -  \\
    % SEED \citep{chen2310seed} & End-to-end  & -   & Linear  & -  & - & - & - \\
    Agent K v1.0  \citep{grosnit2024large} & End-to-end  & -   & Linear  & - & \textcolor{teal}{\ding{52}} & \textcolor{teal}{\ding{55}} \\
    % OpenAI Code Interpreter v2 \citep{openai2024gpt4} & Conversational & System & Linear & \textcolor{teal}{\ding{52}} & \textcolor{teal}{\ding{52}} & \textcolor{red}{\ding{55}} \\
    GPT-4o \citep{openai2024gpt4ocard} & End-to-end & System & - & \textcolor{red}{\ding{55}} & \textcolor{teal}{\ding{52}} & \textcolor{teal}{\ding{52}} \\
    AutoGen Studio \citep{wu2023autogen} & End-to-end & System & Linear & \textcolor{red}{\ding{55}} & \textcolor{teal}{\ding{52}} & \textcolor{teal}{\ding{52}} \\
    Colab Data Science Agent \citep{googleblog2024datascienceagent} & End-to-end & IDE-based & Linear & \textcolor{teal}{\ding{52}} & \textcolor{teal}{\ding{52}} & \textcolor{red}{\ding{55}} \\
    \end{tblr}
    \end{adjustbox}
\end{table}

\subsection{User Interface}

The user interface is crucial for attracting users at first glance. Current research on user interface design can be broadly categorized into four types: Integrated Development Environment-based (IDE-based), Independent System, Command line-based (Command-based), and Operation System-based (OS-based).

\noindent\textbf{IDE-based} \quad Integrated Development Environments (IDEs) such as Jupyter provide convenient tools for data science and analysis. Recent efforts, including Colab Data Science Agent \citep{googleblog2024datascienceagent}, Jupyter-AI \citep{jupyterlab}, Chapyter \citep{chapyter}, and MLCopilot \citep{zhang2023mlcopilot}, have incorporated LLMs into Jupyter environments. For example, Colab Data Science Agent enables planning, automatic code cell generation, execution, and result presentation in the notebook. This approach is particularly popular because it allows users to review, edit, and run code directly.

\noindent\textbf{Independent System} \quad Some works have focused on developing independent systems equipped with user interfaces. For example, ChatGPT introduced a streamlined, intuitive conversational system—a model of interaction that has been widely adopted in subsequent projects. In the context of data analysis tasks, beyond basic text-based input and output, several systems have introduced specialized features, such as visualization, report generation, and file download options, to simplify user interactions. For instance, LAMBDA \citep{sun2024lambda} facilitates easy data review by enabling intuitive data display after users upload their data. Data Formulator 2 \citep{wang2024data} further enhances the iterative process of creating data visualizations through a multi-modal interface, combining graphical user interface (GUI) elements with natural language inputs, allowing users to specify their visualization intentions with both precision and flexibility. WaitGPT \citep{xie2024waitgpt} addresses the challenge of understanding and verifying LLM-generated code by transforming raw code into an interactive, step-by-step visual representation. This allows users to comprehend, validate, and adjust specific data operations, actively guiding and refining the analysis process.

\noindent\textbf{Command Line-based}\quad Works like Data Interpreter \citep{hong2024data} and TaskWeaver \citep{qiao2023taskweaver} using command-line interfaces (CLI) in their works. For researchers and experienced users, it provides greater flexibility and control over the system, allowing users to execute a wide range of functions in the command line and customize their actions. Besides, command-based interfaces often require less computational overhead compared to graphical user interfaces, making them more efficient.

\noindent\textbf{OS-based} \quad OS-based agents, such as UFO \citep{zhang2024ufo}, are designed to operate directly within an operating system environment, allowing them to control a wide range of system tasks and resources. Similarly, Spider2-V \citep{Spider2-V} simulates the typical workflow of a data scientist by mimicking actions such as clicking, typing, and writing code, providing an OS-level interactive experience that closely resembles how humans manage data science tasks. However, while OS-based agents like Spider2-V lay a solid foundation for user interaction, achieving full automation of the data science workflow remains an ongoing challenge \citep{Spider2-V}.

\subsection{Planning, Reasoning, and Reflection}
Planning, Reasoning, and Reflection often play crucial roles in guiding the actions of data agents. In particular, planning and reasoning emphasize the generation of a logically structured sequence or roadmap of actions and thought processes to systematically address problems step by step \citep{huang2024understanding, hong2024data}.
Complex tasks often require a step-by-step approach to ensure effective resolution, while simpler tasks can be handled without such detailed breakdowns. Recently, GPT-4o \citep{openai2024gpt4ocard} introduces a planning architecture that integrates external tools and decomposes complex tasks into structured sub-tasks, enabling more accurate and controllable multi-step reasoning.

Some approaches focus on building conversational data agents \citep{zhang2023data, zhang2023mlcopilot, sun2024lambda}, where users interact with the agent over multiple rounds to complete a task. In these cases, under human supervision, complex planning is not necessary, as guidance can simplify decision-making and adjust the workflow dynamically. Some of these works operate in a Basic I/O mode. On the other hand, End-to-end data agents \citep{guo2024ds, qiao2023taskweaver,hong2024data,chi2024sela,jiang2024aide,li2024autokaggle,trirat2024automl,grosnit2024large} are designed to allow users to issue a single prompt that encompasses all requirements. In these cases, the agent employs planning, reasoning, and reflection to iteratively complete all tasks autonomously.

\begin{figure}
    \centering
    \includegraphics[width=\textwidth]{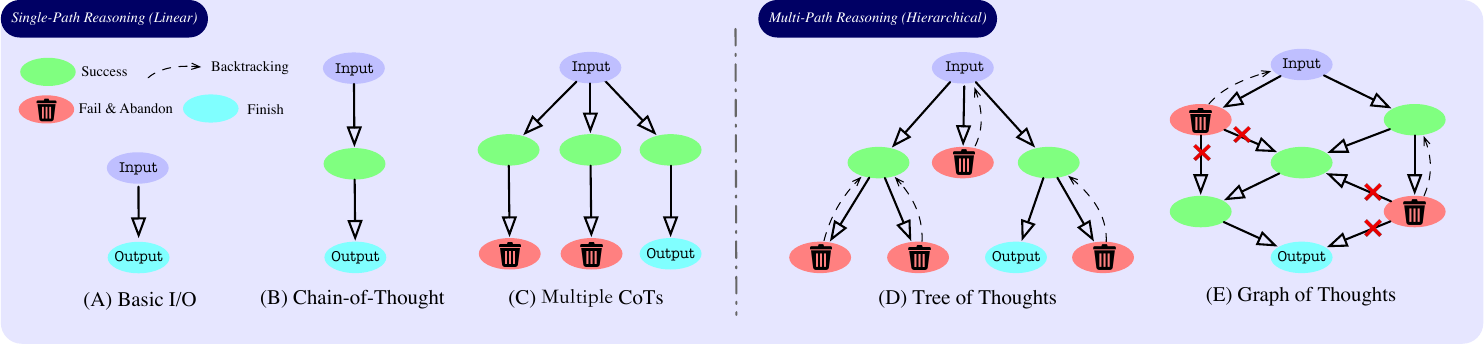}
    \caption{Commonly used planning and reasoning strategies in LLM-based data agents for organizing tasks or solving problems. Each node represents a sub-task in the roadmap.} \label{fig:planning}
\end{figure}

Recent research in planning has introduced two main approaches: Linear Structure Planning (or Single Path Planning/Reasoning) and Hierarchical Structure Planning (or Multiple Path Planning/Reasoning). Figure \ref{fig:planning} illustrates some recent planning methodologies like Chain-of-Thought (CoT) \citep{wei2022chain}, ReAct \citep{yao2022react}, Tree-of-Thoughts (ToT) \citep{yao2024tree}, and Graph-of-Thoughts (GoT) \citep{besta2024graph}.

\noindent\textbf{Linear Structure Planning}\quad In linear structure planning, a task is decomposed into a sequential, step-by-step process. For example, DS-Agent \citep{guo2024ds} utilizes Case-Based Reasoning to retrieve and adapt relevant insights from a knowledge base of past successful Kaggle solutions. This approach allows the agent to learn from previous experiences and continuously improve its performance. Similarly, AutoML-Agent \citep{trirat2024automl} adopts a retrieval-augmented planning (RAP) strategy to generate diverse plans for AutoML tasks. By leveraging the knowledge embedded in LLMs, information retrieved from external APIs, and user requirements, RAP allows the agent to explore a wider range of potential solutions, leading to more optimal plans.

\noindent\textbf{Hierarchical Structure Planning}\quad Simple linear planning is often insufficient for complex tasks. Such tasks may require hierarchical and dynamic, adaptable plans that can account for unexpected issues or errors in execution \citep{hong2024data}. For instance, \cite{hong2024data} utilizes a hierarchical graph modeling approach that breaks down intricate data science problems into manageable sub-problems, represented as nodes in a graph, with their dependencies as edges. This structured representation enables dynamic task management and allows for real-time adjustments to evolving data and requirements. Additionally, they further introduce ``Programmable Node Generation," to automate the generation, refinement, and verification of nodes within the graph, ensuring accurate and robust code generation. AIDE \citep{jiang2024aide} employs Solution Space Tree Search to iteratively improve solutions through generation, evaluation, and selection components. Similarly, SELA \citep{chi2024sela} combines LLMs with Monte Carlo Tree Search (MCTS) to enhance AutoML performance. It starts by using LLMs to generate insights for various machine learning stages, creating a search space for solutions. MCTS then explores this space by iteratively selecting, simulating, and back-propagating feedback, enabling the discovery of optimal pipelines. Agent K v1.0 \citep{grosnit2024large} employs a structured reasoning framework with memory modules, operating through multiple phases. The first phase, automation, handles data preparation and task setup, generating actions through structured reasoning. The second phase, optimization, involves solving tasks and enhancing performance using techniques such as Late-Fusion Model Generation and Bayesian optimization. The final phase, generalization, utilizes a memory-driven system for adaptive task selection.

\noindent\textbf{Reflection}\quad Reflection enables an agent to evaluate past actions and decisions, adjust strategies, and improve future task performance. This process is essential for self-correction and debugging during task execution. For example, \cite{wang2024executable} employs trajectory filtering to train agents that can learn from interactions and enhance their self-debugging capabilities. This technique involves selecting trajectories in which the model initially makes errors but successfully corrects them through self-reflection in subsequent interactions. Similarly, Data-copilot \citep{zhang2023data} and LAMBDA \citep{sun2024lambda} use self-reflection based on code execution feedback to address errors. If a compilation error occurs, the agents repeatedly attempt to revise the code until it runs successfully or a maximum retry limit is reached. This iterative process helps ensure code correctness and usability.

\subsection{Multi-agent Collaboration}
  Multi-agent System (MAS) enable task decomposition through role assignment. In this setup, agents communicate, negotiate, and share information to optimize their collective performance \citep{xi2023rise, liang2024cmat}. It offers several advantages over single-agent setups. First, they reduce redundant and complex context accumulation by isolating responsibilities across agents. Second, each agent instance can be powered by a different language model, opening opportunities to specialize models for domain-specific expertise. For example, in LAMBDA \citep{sun2024lambda}, a dedicated Programmer Agent is responsible for code generation, while noisy error outputs are handled separately by an Inspector Agent. This separation helps the Programmer Agent avoid context overload, simplifies historical trace management, and ultimately improves response accuracy.

  AutoGen introduces a programming framework specifically designed for constructing MAS \citep{wu2023autogen}. Furthermore, AutoML-Agent \citep{trirat2024automl} involves the Agent Manager, Prompt Agent, Operation Agent, Data Agent, and Model Agent—that together cover the entire pipeline, from data retrieval to model deployment. OpenAgents \citep{xie2023openagents} consisted of agents such as the Data Agent, Plugins Agent, and Web Agent. Similarly, AutoKaggle \citep{li2024autokaggle} employs agents like Reader, Planner, Developer, Reviewer, and Summarizer to manage each phase of the process, ensuring comprehensive analysis, effective planning, coding, quality assurance, and detailed reporting. These collaborating mode help decentralized the complicated task, allowing each agent to focus on its specific role, thereby enhancing the overall efficiency and effectiveness of the data analysis process.

\subsection{Knowledge Integration}
Integrating domain-specific knowledge into data agents presents a challenge \citep{dash2022review, sun2024lambda}. For example, when a domain expert has specialized knowledge, such as specific protein analysis code, the agent system are expected able to incorporate and apply this knowledge effectively. One approach is tool-based, where the expert's analysis code is treated as a tool that is recognizable by the LLM \citep{xie2023openagents}. When the agent encounters a relevant problem, it can call upon the appropriate tool from its library to execute the specialized analysis. Another method involves the Retrieval-Augmented Generation (RAG) technique \citep{lewis2020retrieval}, where relevant code is first retrieved and then embedded within the context to facilitate in-context learning. LLM-based agents can also access and interact with external knowledge sources, such as databases or knowledge graphs, to augment their reasoning capabilities \citep{wang2024executable}.

\cite{sun2024lambda} proposes a Knowledge Integration method that builds on this concept. In LAMBDA, analysis codes are parsed into two parts: descriptions and executable code. These are then stored in a knowledge base. When the agent receives a task, it retrieves the relevant knowledge based on the similarity between the task description and the descriptions stored in the knowledge base. The corresponding code is then used for in-context learning (ICL) or back-end execution, depending on the configuration. This approach enables agents to effectively leverage domain-specific knowledge in relevant scenarios.

\subsection{Benchmarks for Evaluating Data Agents}
Evaluating the performance of data agents is crucial for understanding their effectiveness and reliability. Current benchmarks primarily rely on deterministic output comparisons, where an LLM processes a task, generates code, and is evaluated based on the final execution results.
For example, DS-1000 \citep{lai2022ds1000naturalreliablebenchmark} provides a large-scale benchmark of 1000 realistic problems spanning seven core Python data science libraries, with execution-based multi-criteria evaluation and mechanisms to reduce memorization bias. MLAgentBench \citep{huang2024mlagentbenchevaluatinglanguageagents} introduces a benchmark focused on machine learning research workflows by constructing an LLM-agent pipeline. Furthermore, InfiAgent-DABench \citep{hu2024infiagent} presents a end-to-end benchmark for evaluating the capabilities of data agents, the tasks require agents to end-to-end solving complex tasks by interacting with an execution environment. However, for tasks such as data visualization, the outputs are often difficult to compare directly. Designing effective evaluation strategies for data visualizations remains an open and important question.

\subsection{System Design and Other Related Works} %rename the title  Function-specific design | related works

Recent advancements in interactive data science systems highlight a variety of approaches in system design, with LLMs and structured frameworks significantly enhancing the user experience across key areas such as data visualization, task specification, predictive modeling, and data exploration. Notable systems like VIDS \citep{hassan2023chatgpt}, Data-Copilot \citep{zhang2023data}, InsightPilot \citep{ma2023insightpilot}, and JarviX \citep{liu2023jarvix} exemplify diverse design principles tailored to these specific functions. For instance, Data-Copilot adopts a code-centric approach, generating intermediate code to process data and subsequently transforming it into visual outputs, such as charts, tables, and summaries \citep{zhang2023data}.

Other frameworks emphasize workflow automation. InsightPilot integrates an ``insight engine" that guides data exploration, reducing LLM hallucinations and enhancing the accuracy of exploratory tasks \citep{ma2023insightpilot}. JarviX, in combination with MLCopilot \citep{zhang2023mlcopilot}, contributes to automated machine learning by merging LLM-driven insights with AutoML pipelines. Additionally, in the domain of database management, systems like LLMDB \citep{zhou2024llm} improve efficiency and reduce hallucinations and computational costs during tasks such as query rewriting, database diagnosis, and data analytics. In terms of data visualization, MatPlotAgent \citep{yang2024matplotagent} transforms raw data into clear, informative visualizations by leveraging both code-based and multi-modal LLMs.

Moreover, Data Formulator 2 \citep{wang2024data} organizes user interactions into "data threads" to provide context and facilitate the exploration and revision of prior steps. A similar approach is seen in WaitGPT \citep{xie2024waitgpt}, which transforms raw code into an interactive visual representation. This provides a step-by-step visualization of LLM-generated code in real-time, allowing users to understand, verify, and modify individual data operations. SEED \citep{chen2310seed} combines LLMs with methods like code generation and small models to produce domain-specific data curation solutions. HuggingGPT \citep{shen2024hugginggpt}, on the other hand, uses LLMs to coordinate a variety of expert models from platforms such as Hugging Face, solving a broader range of AI tasks across multiple modalities.

Lastly, in terms of industry applications, lots of companies have used agents in the business analysis. For example FUTU use AI to analyze the stock market and provide investment advice \citep{futu}. Julius \citep{julius} facilitates data science education by building a bridge that allowing professors to create interactive workflows for lessons, which can be shared with students for a seamless teaching experience through natural language interaction.

\section{Data Analysis Through Natural Language Interaction: Case Studies}\label{sec:case_study}

In this section, we present a series of case studies conducted by a diverse range of agents, each illustrating the new data analysis paradigm facilitated through natural language interaction. These case studies demonstrate how this approach enables users to engage with data more intuitively and effectively, breaking down traditional barriers to data accessibility and understanding. By leveraging natural language processing, these agents can interpret and respond to complex queries, providing insights that are both comprehensive and easily digestible. Through these examples, we aim to highlight the transformative potential of natural language interaction in data analysis.

\subsection{Case study 1: Exploratory Data Analysis and %Machine Learning
Model Building by Conversational Data Agents}
In this case study, we utilized ChatGPT and LAMBDA to demonstrate exploratory data analysis (EDA) and a simple model building process.
Specifically, we first used ChatGPT to explore the effect of alcohol content on the quality of different types of wine, focusing on both red and white varieties. Then, we used LAMBDA to illustrate an interactive modeling process and automatically generate analysis reports.

\begin{figure}[t]
        \centering
        \includegraphics[width=\textwidth]{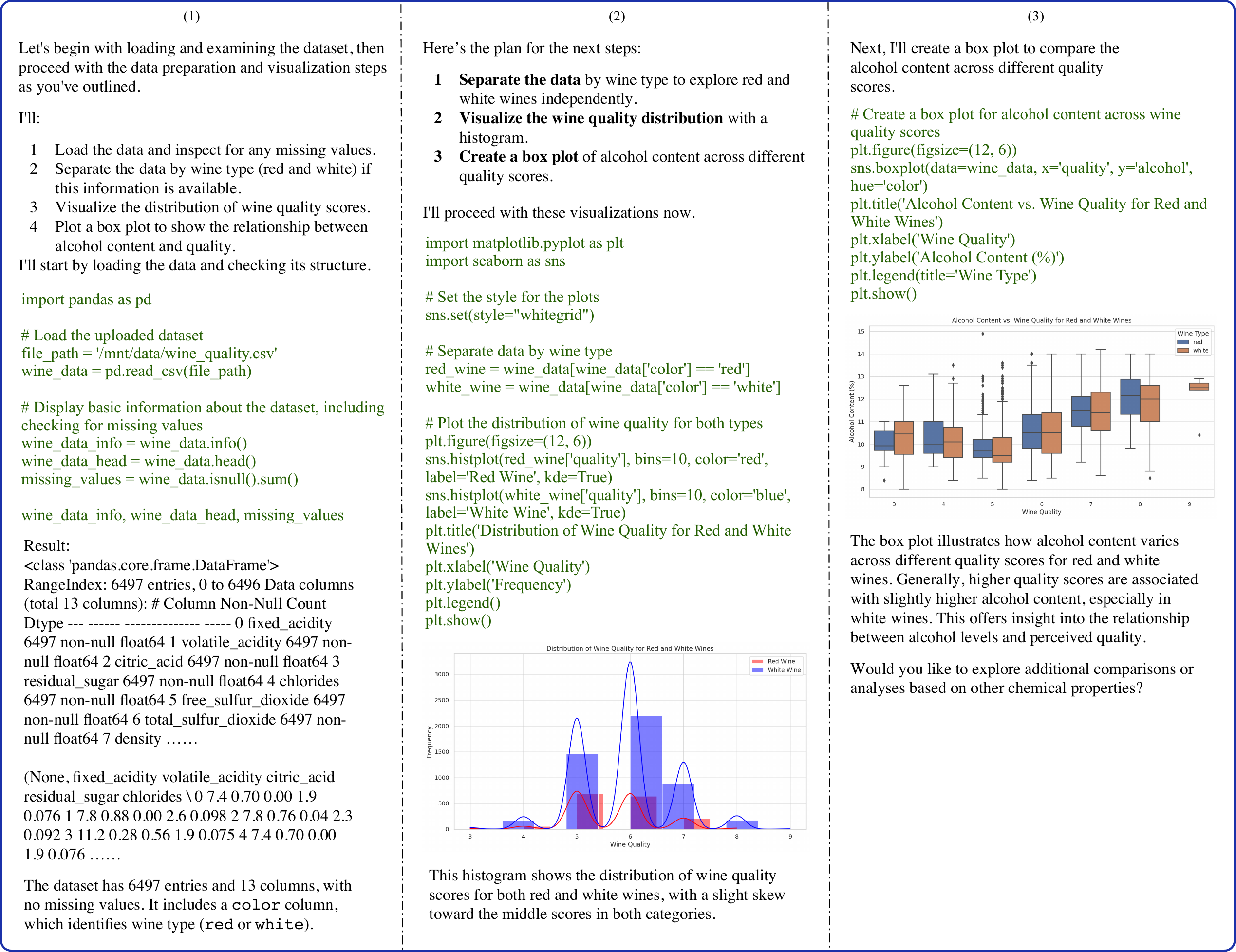}
        \caption{Partial dialogue from the ChatGPT-Advanced Data Analysis in Case Study 1.  Items 1-4 list the work done by ChatGPT in each step.}
        \label{fig:case1}
\end{figure}

We used the Wine Quality dataset, a tabular dataset with dimension $4898 \times 11$. The goal is to examine how 10 coviarates in this dataset affect the wine quality rating. We employed ChatGPT-ADA to conduct EDA and visualize the influence of alcohol content on wine quality ratings. Figure \ref{fig:case1} illustrates the detailed planning and problem-solving process.

GPT-ADA first analyzed the problems and then outlined a step-by-step plan to solve the tasks. The entire workflow proceeded smoothly, with the code running efficiently to load the data, check for missing values, and generate visualizations, with each step delivering accurate results. Its ability to interpret data and provide insights significantly streamlined the analytical process. Finally, it provided insights into the relationship between quality scores and alcohol content.

\begin{figure}[t]
    \centering
    \includegraphics[width=\textwidth]{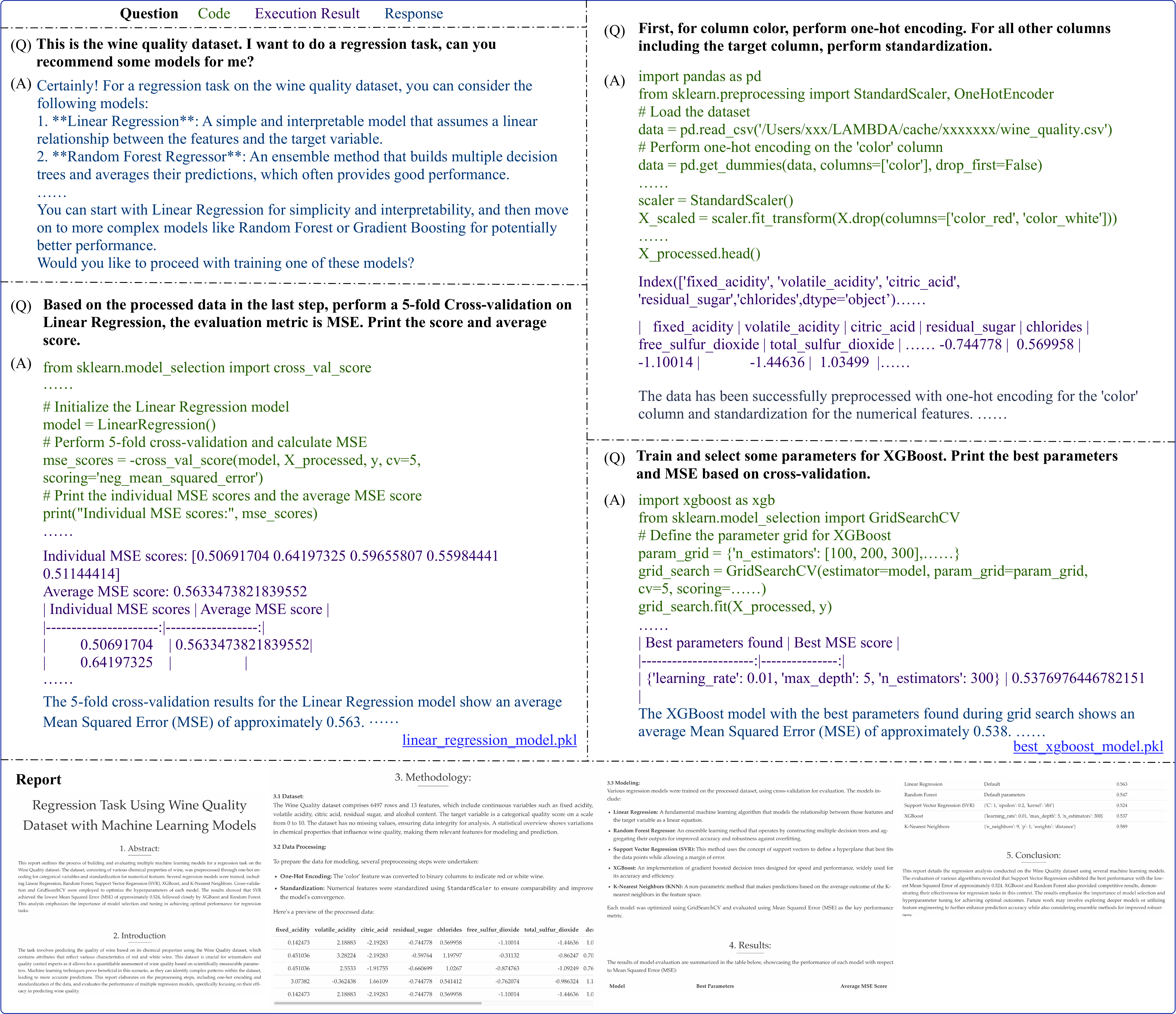}
    \caption{Conversational machine learning and report generation by LAMBDA. Excerpt from a partial dialogue.}
    \label{fig:case_lambda}
\end{figure}

Next, we train a set of models to predict wine quality using LAMBDA. LAMBDA facilitates an interactive analysis process, enabling us to perform tasks such as data processing, feature engineering, model training, parameter tuning, and evaluation through a series of guided conversations. Finally, we used LAMBDA’s built-in report generation feature to compile an analysis report, which includes details of the tasks completed in the conversation history. The analysis process, including the conversation and the generated report, is presented in Figure \ref{fig:case_lambda}.

As beginner-level users, we first asked LAMBDA to recommend some models, and it suggested advanced options like XGBoost. Next, we tasked LAMBDA with basic data preprocessing, which it handled correctly. We then trained and evaluated the recommended models using 5-fold cross-validation, a task LAMBDA performed exceptionally well, even providing download links for the resulting models. Finally, we used LAMBDA’s report generation feature to create a structured and comprehensive report that effectively captured the key insights.

This example demonstrates the effectiveness of conversational data agents like ChatGPT and LAMBDA in streamlining the data visualization and machine learning workflow, particularly for users without programming experience.

\subsection{Case Study 2: Residual Diagnostics and Heteroskedasticity Testing}

To examine the ability of LLM-based data agents to perform statistically rigorous regression diagnostics, we prompted LAMBDA and GPT-4o to conduct a linear regression analysis using the Auto MPG dataset, a tabular data with dimension of 398 $times$ 7. The goal was to predict \texttt{mpg} (miles per gallon) based on vehicle characteristics, notably \texttt{horsepower} and \texttt{weight}. The prompt and response of LAMBDA are detailed in the figure \ref{fig:cs_stat}.

LAMBDA correctly loaded the dataset, performed appropriate preprocessing (e.g., handling non-numeric entries), and fit a linear model using \texttt{statsmodels}. It then computed and visualized residuals, followed by executing the Breusch–Pagan test for heteroskedasticity. The test output included the LM statistic and associated p-value, indicating a strong violation of the homoskedasticity assumption.

The residual plot visually confirmed increasing residual variance with larger fitted values. LAMBDA also summarized next steps, suggesting robust standard errors or model transformation to address heteroskedasticity. This example demonstrates LAMBDA’s ability to execute, interpret, and communicate statistically meaningful diagnostics in a flexible code-first environment. Besides, GPT-4o was also able to complete the same task successfully; further details and chat transcripts can be found in the supplementary materials.

\subsection{Case Study 3: Bootstrap Confidence Intervals}

\begin{figure}[h]
  \centering
  \includegraphics[width=\textwidth]{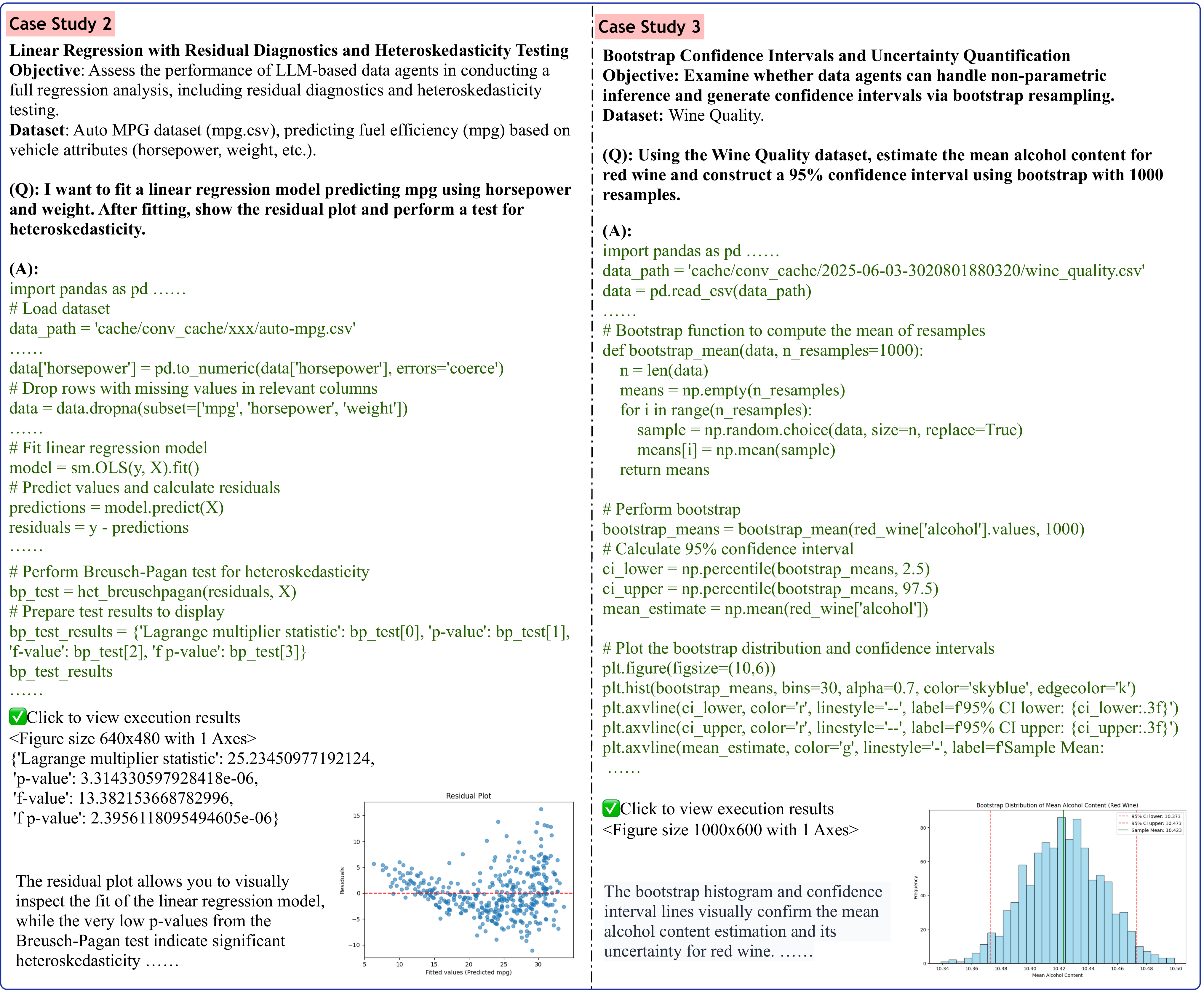}
  \caption{Partial dialogue from residual diagnostics and heteroskedasticity testing, and bootstrap confidence interval estimation.}
  \label{fig:cs_stat}
\end{figure}

In this case study, we assessed whether LLM-based data agents can perform non-parametric inference through bootstrap resampling. Using the Wine Quality dataset, the task was to estimate the average alcohol content for red wine and construct a 95\% confidence interval using 1000 bootstrap resamples. Figure~\ref{fig:cs_stat} shows the interaction with LAMBDA for completing this task.

LAMBDA successfully filtered the dataset to isolate red wines, extracted the \texttt{alcohol} variable, and implemented the bootstrap routine by repeatedly sampling with replacement. It then computed the empirical 2.5th and 97.5th percentiles of the bootstrapped means to form the confidence interval. The agent also produced a histogram showing the bootstrap distribution, overlaid with the CI bounds and sample mean.

This case illustrates that LAMBDA is capable of performing robust uncertainty quantification and generating high-quality visual explanations without relying on strict parametric assumptions. GPT-4o also successfully completed this task; its outputs and detailed interactions are included in the supplementary materials.

We found that different prompting may lead to differences in implementation details, such as the choice of hyperparameters or types of plots.

\subsection{Case study 4: Expandability of Data Agents}
In many situations, we encounter tasks that cannot be handled effectively using LLMs because their training data do not include the necessary knowledge for such tasks. In these cases, if a data agent is designed to be extensible, manual tool expansion or knowledge integration can address this limitation. In this case study, we demonstrate how both the Data Interpreter and LAMBDA leverage integration mechanisms to incorporate additional packages or domain-specific knowledge.

\begin{figure}[t]
\centering
\includegraphics[width=\textwidth]{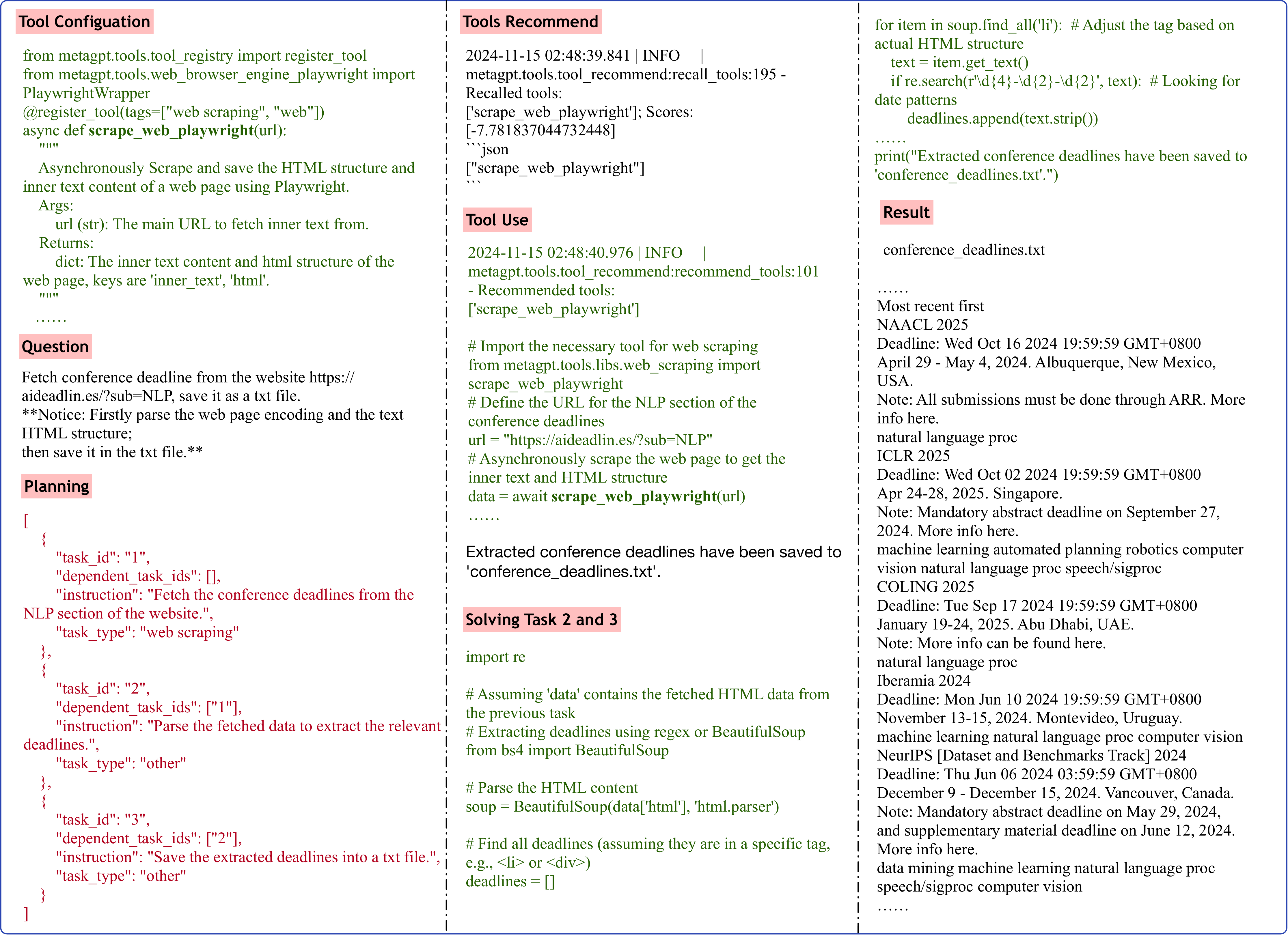}
\caption{Creating and using the customized tool in the Data Interpreter. Excerpt from a partial dialogue.} \label{fig:tool}

\end{figure}

\noindent\textbf{Tools Integration in Data Interpreter}\ In this example, our objective is to extract submission deadlines for AI conferences from a public website\footnote{\url{https://aideadlin.es}} and save the results. We prompted the agent with the target URL and the desired output format. The agent successfully identified relevant information such as conference names and deadlines and generated structured output. The complete workflow, including prompt, execution, and results, is shown in Figure~\ref{fig:tool}.

In this example, the Data Interpreter began with an initial plan. For each sub-task, it recommended relevant tools with a score indicating their suitability. The system then decided whether to use the suggested tool. For instance, it used \texttt{scrape\_web\_playwright} for a web-scraping task. This iterative recommendation and tool selection process continued until all sub-tasks were completed, addressing limitations in LLMs' built-in abilities and knowledge.

\noindent\textbf{Knowledge Integration in LAMBDA}\ In this example, we consider the problem of training a  Fixed Point Non-Negative Neural Network (FPNNN), which is defined as a neural network that maps nonnegative vectors to nonnegative vectors.
We train a FPNNN with MNIST data. First, we integrated the code into the knowledge base. Then, we defined the model as \texttt{Core} and delineated the \texttt{Core} function, which directly accepts parameters, and the \texttt{Runnable} function, which was defined and executed separately. Figure \ref{fig:case_lambda} presents the configuration, prompt, and problem-solving process.

\begin{figure}[t]
\centering
\includegraphics[width=\textwidth]{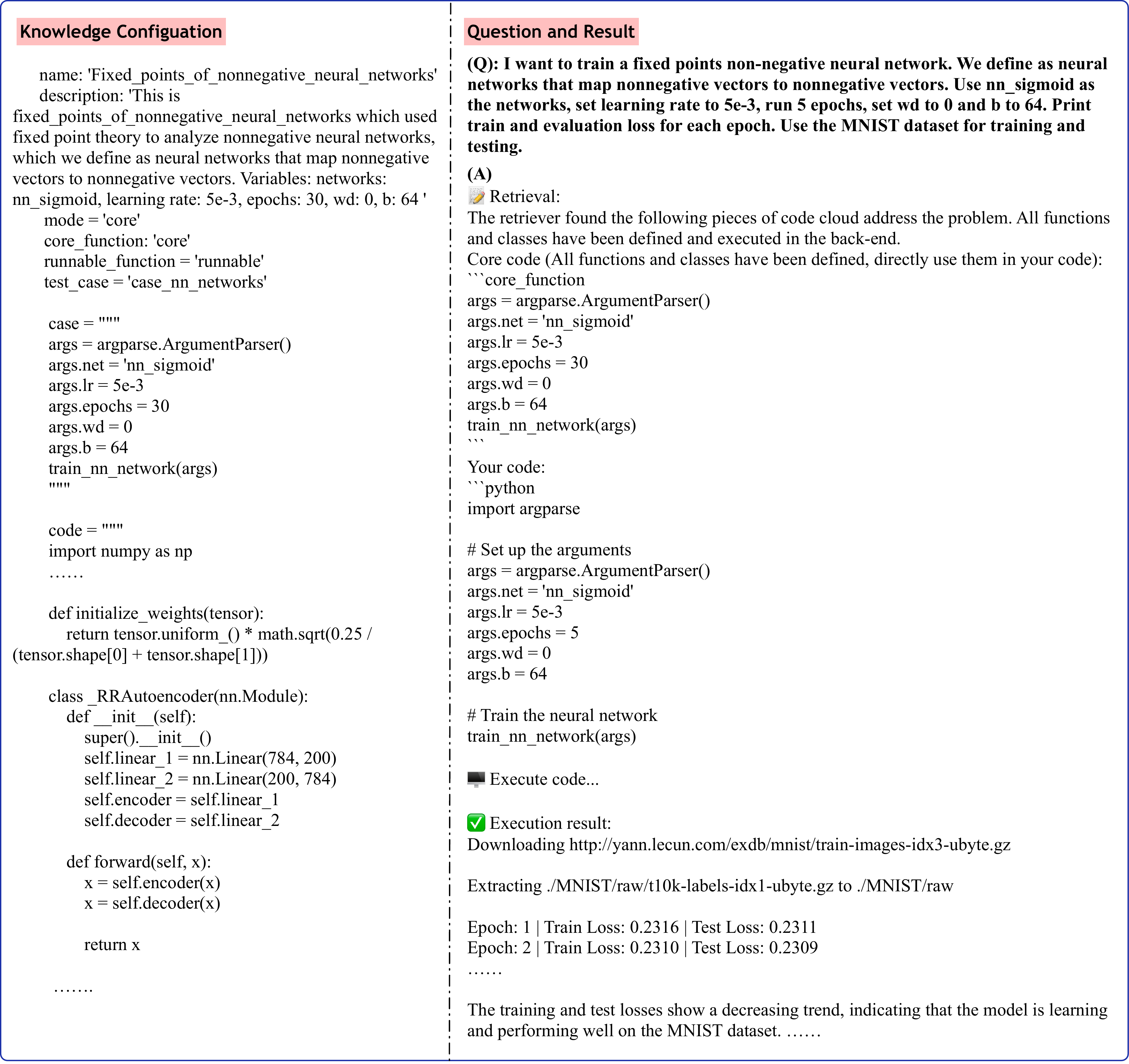}
\caption{Integrating knowledge of FPNNNs in LAMBDA. Excerpt from a partial dialogue.} \label{fig:knw}
\end{figure}

LAMBDA first retrieved the relevant code from the knowledge base, and then its \texttt{Core} function was presented in the context. By modifying the core code, LAMBDA generated the correct code and completed the task successfully.

\section{Challenges and Future Directions}\label{sec:future}
In this section, we highlight some challenges and suggest future directions in using LLMs or LLM-based data agents for statistical analysis.
%We discuss the challenges from the following

\subsection{Challenges in the Capabilities of LLMs}

LLMs function as the ``brain" of a data agent, interpreting user intent and generating structured plans to carry out data analysis tasks. For a data agent to be effective, it must possess advanced knowledge in statistics, data science, and programming, enabling it to support users throughout the analytical process.

\noindent\textbf{Advanced Models} \quad Current state-of-the-art models like GPT-4 show strong performance on undergraduate-level mathematics and statistics problems, yet struggle with more advanced, graduate-level tasks \citep{frieder2023mathematicalcapabilitieschatgpt}. Additionally, the success rate of fully automating complete data workflows with current agents remains low \citep{Spider2-V}. This suggests that enhancements in LLMs, particularly in knowledge of statistics and data analysis, are still needed.

\noindent\textbf{Multi-Modality and Reasoning} \
A key challenge for current LLMs lies in processing multi-modal inputs, including charts, tables, and code, which are essential to data analysis workflows \citep{inala2024data}. Future advancements may improve the ability to perform reasoning across mixed modalities, such as generating visualizations by replicating the style of an input visualization. Besides, data agent can further integrate reinforcement learning for enhanced code generation \citep{wang2024enhancing} and human-centric reward optimization \citep{zhou2024human} while also incorporating methods to improve the coherence and retrieval capabilities of its analytical outputs \cite{yi2025score}s's.

\subsection{Challenges in Statistical Analysis}
\noindent\textbf{Intelligent Statistical Analysis Software} \
While established tools such as SPSS and R are highly mature, data agents have the potential to transform statistical analysis through intelligent assistance. To realize this vision, agents must support flexible package integration, facilitate contributions from domain experts, and remain aligned with evolving programming ecosystems. Such a collaborative framework could accelerate innovation in the field. Furthermore, by guiding users and recommending appropriate methods, data agents can enhance research efficiency and expand access to advanced statistical techniques.

\noindent\textbf{Incorporating Other Large Models into Statistical Analysis} \
Statistical analysis of complex data is increasingly leveraging representations generated by large models for research purposes. For example, in predicting the tertiary structure of proteins, LLMs can utilize representations of primary and secondary structures—capabilities that traditional statistical software such as Matlab and R currently lack. Similarly, in the analysis of electronic health records, LLMs are being used to construct meaningful representations that facilitate downstream analysis. If data agents can effectively harness domain-specific knowledge models, they have the potential to significantly advance statistical and data science research, enabling more sophisticated analyses and fostering deeper insights across scientific disciplines.

\subsection{Challenges in Real-World Adoption}
Although the data agents have shown great potential in improving the accessibility of data analysis, there are still several challenges that need to be addressed for real-world adoption.

\noindent\textbf{Trade-off Between Hardware and Privacy} \
First, deploying large language models often requires high-performance computing resources. Running these models on CPU-only machines results in slow inference. API-based solutions also raise concerns about data privacy and security, as sensitive information may be transmitted to external servers. This is especially critical in fields such as healthcare and finance, where data confidentiality is paramount. Therefore, developing lightweight, expert-level data science models that can run efficiently on local machines without compromising performance is essential.

\noindent\textbf{High-concurrency System} \ High-concurrency environments pose significant scalability issues. In client-server architectures where each user session is associated with an isolated sandbox for secure code execution, the server may experience substantial resource strain under heavy load. Maintaining a large number of concurrent sandboxes can overwhelm system resources, leading to degraded performance or system instability. Therefore, the design of efficient scheduling algorithms to manage limited computational resources across multiple sandbox instances becomes critical.

\noindent\textbf{Integration with Existing Workflows} \ While data agents excel in lowering the barrier to entry for non-programmers, they currently lack the flexibility and debugging capabilities of traditional IDEs. This makes them less suitable for complex, customized workflows that require iterative development and fine-grained control. A promising direction is to support the seamless export of an agent’s actions \citep{sun2024lambda}, such as executed code, into IDEs like Jupyter Notebooks, which can serve as a bridge for smoother integration with conventional tools and workflows.

\section{Discussion}\label{sec:diss}

\subsection{Model Level Reproducibility}
\noindent While data agents are generally robust to variations in prompt phrasing and can reliably complete the intended analytical tasks, we observed notable differences in their reasoning processes and implementation details. For example, when prompted to perform regression diagnostics, different phrasings such as “analyze residuals” versus “check model assumptions” resulted in the same core analysis but with different statistical tests or plotting choices. Similarly, in visualization tasks, one prompt might produce a bar chart while another yields a pie chart, depending on how the goal is described. Even for model training, default hyperparameters, such as learning rate or number of iterations, could vary slightly across prompts, leading to differences in performance metrics. These variations do not typically prevent task completion but can impact result interpretability, especially in rigorous statistical workflows where consistency across runs is critical.

\subsection{System Level Reproducibility}
\noindent \textbf{Experiment Setting} \
Experiment reproducibility can be enhanced through careful experiment designs. For example, LAMBDA \citep{sun2024lambda} incorporates built-in mechanisms to export the full execution history into executable formats such as Jupyter Notebooks. When combined with proper experiment controls, such as setting random seeds, these exports enable end-to-end reproducibility of experimental results.
In addition, designing human-in-the-loop mechanisms allows users to inspect, edit, or revise the code generated by LLMs during the problem-solving process. This interactive approach further supports reproducibility by enabling manual correction and verification of intermediate steps.

\noindent \textbf{Version Control and Workflow Management} \
Version control tools such as Git can enhance reproducibility by tracking changes in code, data, and prompts, making it easier to reproduce results and collaborate with others. Furthermore, workflow management systems like Snakemake and Nextflow allow users to define and automate each step of the analysis pipeline, ensuring that processes can be reliably repeated. When used alongside data agents, these tools can greatly improve both reproducibility and transparency. However, most current data agents lack native support for these tools, presenting opportunities for future development.

\section{Conclusion}\label{sec:con}
This survey has
explored the recent progress of LLM-based data science agents. These agents have shown great potential in making data analysis more accessible to a wider range of users, even those with limited technical skills. By leveraging the capabilities of LLMs, they are able to handle various data analysis tasks, from data visualization to machine learning, through natural language interaction.

However, as discussed, they also face several challenges. In terms of model capabilities, improvements are needed in domain-specific knowledge and multi-modal handling. For intelligent statistical analysis software, seamless package management and community building are crucial. Additionally, effectively integrating other large models into statistical analysis and addressing data infrastructure and evaluation issues remain important areas for future development.

Overall, while LLM-based data science agents have made significant strides, continuous research and innovation are required to overcome the existing challenges and fully realize their potential in revolutionizing the field of data analysis.

\subsection*{Acknowledgments}
The authors are grateful to the Editor, Associate Editor and two anonymous reviewers
%and the reproducibility reviewer
for their valuable comments and suggestions, which significantly improved the quality of the paper.

\subsection*{Funding}
This work was funded by the Centre for the Mathematical Foundations of Generative AI
and the research grants from The Hong Kong Polytechnic University (P0046811).
The research of Ruijian Han was partially supported by The Hong Kong RGC grant (14301821) and The Hong Kong Polytechnic University (P0044617, P0045351, P0050935). The research of Binyan Jiang was partially supported by The Hong Kong RGC grant (15302722).
The research of Houduo Qi was partially supported by the Hong Kong RGC grant (15309223) and
The Hong Kong Polytechnic University (P0045347).
The research of Defeng Sun and Yancheng Yuan was partially supported by the Research Center for Intelligent Operations Research at The Hong Kong Polytechnic University (P0051214).
The research of Jian Huang was partially supported by The Hong Kong Polytechnic University (P0042888, P0045417, P0045931).

\subsection*{Disclosure Statement}
The authors report there are no competing interests to declare.

{\singlespace
\bibliographystyle{apalike}
\bibliography{LLMDAsurvey_TAS_Final.bib}

\begin{thebibliography}{}

\bibitem[Besta et~al., 2024]{besta2024graph}
Besta, M., Blach, N., Kubicek, A., Gerstenberger, R., Podstawski, M., Gianinazzi, L., Gajda, J., Lehmann, T., Niewiadomski, H., Nyczyk, P., et~al. (2024).
\newblock Graph of thoughts: Solving elaborate problems with large language models.
\newblock In {\em Proceedings of the AAAI Conference on Artificial Intelligence}, volume~38, pages 17682--17690.

\bibitem[Cao, 2017]{cao2017data}
Cao, L. (2017).
\newblock Data science: Challenges and directions.
\newblock {\em Communications of the ACM}, 60(8):59--68.

\bibitem[Cao et~al., 2024]{Spider2-V}
Cao, R., Lei, F., Wu, H., Chen, J., Fu, Y., Gao, H., Xiong, X., Zhang, H., Mao, Y., Hu, W., Xie, T., Xu, H., Zhang, D., Wang, S., Sun, R., Yin, P., Xiong, C., Ni, A., Liu, Q., Zhong, V., Chen, L., Yu, K., and Yu, T. (2024).
\newblock Spider2-v: How far are multimodal agents from automating data science and engineering workflows?

\bibitem[Chandel et~al., 2022]{chandel2022training}
Chandel, S., Clement, C.~B., Serrato, G., and Sundaresan, N. (2022).
\newblock Training and evaluating a jupyter notebook data science assistant.
\newblock {\em arXiv preprint arXiv:2201.12901}.

\bibitem[{chapyter}, 2023]{chapyter}
{chapyter} (2023).
\newblock Chapyter.
\newblock \url{https://github.com/chapyter/chapyter}.

\bibitem[Chen et~al., 2012]{chen2012business}
Chen, H., Chiang, R.~H., and Storey, V.~C. (2012).
\newblock Business intelligence and analytics: From big data to big impact.
\newblock {\em MIS quarterly}, 36(4):1165--1188.

\bibitem[Chen et~al., 2024]{chen2310seed}
Chen, Z., Cao, L., Madden, S., Kraska, T., Shang, Z., Fan, J., Tang, N., Gu, Z., Liu, C., and Cafarella, M. (2024).
\newblock Seed: Domain-specific data curation with large language models. arxiv 2023.
\newblock {\em arXiv preprint arXiv:2310.00749}.

\bibitem[Cheng et~al., 2023]{cheng2023gpt}
Cheng, L., Li, X., and Bing, L. (2023).
\newblock Is gpt-4 a good data analyst?
\newblock {\em arXiv preprint arXiv:2305.15038}.

\bibitem[Cheng et~al., 2024]{cheng2024exploring}
Cheng, Y., Zhang, C., Zhang, Z., Meng, X., Hong, S., Li, W., Wang, Z., Wang, Z., Yin, F., Zhao, J., et~al. (2024).
\newblock Exploring large language model based intelligent agents: Definitions, methods, and prospects.
\newblock {\em arXiv preprint arXiv:2401.03428}.

\bibitem[Chi et~al., 2024]{chi2024sela}
Chi, Y., Lin, Y., Hong, S., Pan, D., Fei, Y., Mei, G., Liu, B., Pang, T., Kwok, J., Zhang, C., et~al. (2024).
\newblock Sela: Tree-search enhanced llm agents for automated machine learning.
\newblock {\em arXiv preprint arXiv:2410.17238}.

\bibitem[Dash et~al., 2022]{dash2022review}
Dash, T., Chitlangia, S., Ahuja, A., and Srinivasan, A. (2022).
\newblock A review of some techniques for inclusion of domain-knowledge into deep neural networks.
\newblock {\em Scientific Reports}, 12(1):1040.

\bibitem[Dong and Wang, 2024]{10.1145/3626772.3661384}
Dong, H. and Wang, Z. (2024).
\newblock Large language models for tabular data: Progresses and future directions.
\newblock In {\em Proceedings of the 47th International ACM SIGIR Conference on Research and Development in Information Retrieval}, SIGIR '24, page 2997–3000, New York, NY, USA. Association for Computing Machinery.

\bibitem[for Statistical~Computing, 1995]{R}
for Statistical~Computing, R.~F. (1995).
\newblock {\em R: A Language and Environment for Statistical Computing}.

\bibitem[Foundation, 1991]{Python}
Foundation, P.~S. (1991).
\newblock {\em Python Programming Language}.

\bibitem[Frieder et~al., 2023]{frieder2023mathematicalcapabilitieschatgpt}
Frieder, S., Pinchetti, L., Chevalier, A., Griffiths, R.-R., Salvatori, T., Lukasiewicz, T., Petersen, P.~C., and Berner, J. (2023).
\newblock Mathematical capabilities of chatgpt.

\bibitem[FUTU, 2024]{futu}
FUTU (2024).
\newblock Futubull ai.

\bibitem[GLM, 2024]{glm2024chatglm}
GLM, T. (2024).
\newblock Chatglm: A family of large language models from glm-130b to glm-4 all tools.

\bibitem[Google, 2025]{googleblog2024datascienceagent}
Google (2025).
\newblock Data science agent in colab with gemini.
\newblock \url{https://developers.googleblog.com/en/data-science-agent-in-colab-with-gemini/}.
\newblock Accessed: 2025.

\bibitem[Grosnit et~al., 2024]{grosnit2024large}
Grosnit, A., Maraval, A., Doran, J., Paolo, G., Thomas, A., Beevi, R. S. H.~N., Gonzalez, J., Khandelwal, K., Iacobacci, I., Benechehab, A., et~al. (2024).
\newblock Large language models orchestrating structured reasoning achieve kaggle grandmaster level.
\newblock {\em arXiv preprint arXiv:2411.03562}.

\bibitem[Guo et~al., 2024]{guo2024ds}
Guo, S., Deng, C., Wen, Y., Chen, H., Chang, Y., and Wang, J. (2024).
\newblock Ds-agent: Automated data science by empowering large language models with case-based reasoning.
\newblock {\em arXiv preprint arXiv:2402.17453}.

\bibitem[Hassan et~al., 2023]{hassan2023chatgpt}
Hassan, M.~M., Knipper, A., and Santu, S. K.~K. (2023).
\newblock Chatgpt as your personal data scientist.
\newblock {\em arXiv preprint arXiv:2305.13657}.

\bibitem[Hong et~al., 2024]{hong2024data}
Hong, S., Lin, Y., Liu, B., Wu, B., Li, D., Chen, J., Zhang, J., Wang, J., Zhang, L., Zhuge, M., et~al. (2024).
\newblock Data interpreter: An llm agent for data science.
\newblock {\em arXiv preprint arXiv:2402.18679}.

\bibitem[Hu et~al., 2024]{hu2024infiagent}
Hu, X., Zhao, Z., Wei, S., Chai, Z., Ma, Q., Wang, G., Wang, X., Su, J., Xu, J., Zhu, M., et~al. (2024).
\newblock Infiagent-dabench: Evaluating agents on data analysis tasks.
\newblock {\em arXiv preprint arXiv:2401.05507}.

\bibitem[Huang et~al., 2024a]{huang2024mlagentbenchevaluatinglanguageagents}
Huang, Q., Vora, J., Liang, P., and Leskovec, J. (2024a).
\newblock Mlagentbench: Evaluating language agents on machine learning experimentation.

\bibitem[Huang et~al., 2024b]{huang2024understanding}
Huang, X., Liu, W., Chen, X., Wang, X., Wang, H., Lian, D., Wang, Y., Tang, R., and Chen, E. (2024b).
\newblock Understanding the planning of llm agents: A survey.
\newblock {\em arXiv preprint arXiv:2402.02716}.

\bibitem[IBM, 1968]{SPSS}
IBM (1968).
\newblock {\em SPSS Statistics}.

\bibitem[Inala et~al., 2024]{inala2024data}
Inala, J.~P., Wang, C., Drucker, S., Ramos, G., Dibia, V., Riche, N., Brown, D., Marshall, D., and Gao, J. (2024).
\newblock Data analysis in the era of generative ai.
\newblock {\em arXiv preprint arXiv:2409.18475}.

\bibitem[Inc., 1976]{SAS}
Inc., S.~I. (1976).
\newblock {\em SAS Software}.

\bibitem[Institute, 2011]{mckinsey2011big}
Institute, M.~G. (2011).
\newblock {\em Big data: The next frontier for innovation, competition, and productivity}.
\newblock McKinsey \& Company.

\bibitem[Jiang et~al., 2024]{jiang2024aide}
Jiang, Z. et~al. (2024).
\newblock {AIDE: the Machine Learning CodeGen Agent}.
\newblock \url{https://github.com/WecoAI/aideml}.
\newblock Accessed: 2024-08-29.

\bibitem[Jordan and Mitchell, 2015]{jordan2015machine}
Jordan, M.~I. and Mitchell, T.~M. (2015).
\newblock Machine learning: Trends, perspectives, and prospects.
\newblock {\em Science}, 349(6245):255--260.

\bibitem[Julius, 2025]{julius}
Julius (2025).
\newblock Julius ai.

\bibitem[{jupyterlab}, 2023]{jupyterlab}
{jupyterlab} (2023).
\newblock Jupyter-ai.
\newblock \url{https://github.com/jupyterlab/jupyter-ai}.

\bibitem[Kitchin, 2014]{kitchin2014big}
Kitchin, R. (2014).
\newblock {\em The data revolution: Big data, open data, data infrastructures and their consequences}.
\newblock Sage.

\bibitem[Lai et~al., 2022]{lai2022ds1000naturalreliablebenchmark}
Lai, Y., Li, C., Wang, Y., Zhang, T., Zhong, R., Zettlemoyer, L., tau Yih, S.~W., Fried, D., Wang, S., and Yu, T. (2022).
\newblock Ds-1000: A natural and reliable benchmark for data science code generation.

\bibitem[Lewis et~al., 2020]{lewis2020retrieval}
Lewis, P., Perez, E., Piktus, A., Petroni, F., Karpukhin, V., Goyal, N., K{\"u}ttler, H., Lewis, M., Yih, W.-t., Rockt{\"a}schel, T., et~al. (2020).
\newblock Retrieval-augmented generation for knowledge-intensive nlp tasks.
\newblock {\em Advances in Neural Information Processing Systems}, 33:9459--9474.

\bibitem[Li et~al., 2024]{li2024autokaggle}
Li, Z., Zang, Q., Ma, D., Guo, J., Zheng, T., Niu, X., Yue, X., Wang, Y., Yang, J., Liu, J., et~al. (2024).
\newblock Autokaggle: A multi-agent framework for autonomous data science competitions.
\newblock {\em arXiv preprint arXiv:2410.20424}.

\bibitem[Liang et~al., 2024]{liang2024cmat}
Liang, X., He, Y., Tao, M., Xia, Y., Wang, J., Shi, T., Wang, J., and Yang, J. (2024).
\newblock Cmat: A multi-agent collaboration tuning framework for enhancing small language models.
\newblock {\em arXiv preprint arXiv:2404.01663}.

\bibitem[Liu et~al., 2023]{liu2023jarvix}
Liu, S.-C., Wang, S., Chang, T., Lin, W., Hsiung, C.-W., Hsieh, Y.-C., Cheng, Y.-P., Luo, S.-H., and Zhang, J. (2023).
\newblock Jarvix: A llm no code platform for tabular data analysis and optimization.
\newblock In {\em Proceedings of the 2023 Conference on Empirical Methods in Natural Language Processing: Industry Track}, pages 622--630.

\bibitem[Luo et~al., 2024]{luo2024autom3l}
Luo, D., Feng, C., Nong, Y., and Shen, Y. (2024).
\newblock Autom3l: An automated multimodal machine learning framework with large language models.
\newblock In {\em Proceedings of the 32nd ACM International Conference on Multimedia}, pages 8586--8594.

\bibitem[Ma et~al., 2023]{ma2023insightpilot}
Ma, P., Ding, R., Wang, S., Han, S., and Zhang, D. (2023).
\newblock Insightpilot: An llm-empowered automated data exploration system.
\newblock In {\em Proceedings of the 2023 Conference on Empirical Methods in Natural Language Processing: System Demonstrations}, pages 346--352.

\bibitem[MathWorks, 1984]{Matlab}
MathWorks (1984).
\newblock {\em MATLAB}.

\bibitem[Microsoft, 1985]{Excel}
Microsoft (1985).
\newblock {\em Microsoft Excel}.

\bibitem[Microsoft, 2013]{PowerBI}
Microsoft (2013).
\newblock {\em Power BI}.

\bibitem[Nejjar et~al., 2023]{nejjar2023llms}
Nejjar, M., Zacharias, L., Stiehle, F., and Weber, I. (2023).
\newblock Llms for science: Usage for code generation and data analysis.
\newblock {\em Journal of Software: Evolution and Process}, page e2723.

\bibitem[OpenAI, 2023]{openai2024gpt4}
OpenAI (2023).
\newblock Gpt-4 technical report.
\newblock {\em arXiv preprint arXiv:2303.08774}.

\bibitem[OpenAI, 2024]{openai2024gpt4ocard}
OpenAI (2024).
\newblock Gpt-4o system card.

\bibitem[Provost and Fawcett, 2013]{provost2013data}
Provost, F. and Fawcett, T. (2013).
\newblock Data science and its relationship to big data and data-driven decision making.
\newblock {\em Big data}, 1(1):51--59.

\bibitem[Qiao et~al., 2023]{qiao2023taskweaver}
Qiao, B., Li, L., Zhang, X., He, S., Kang, Y., Zhang, C., Yang, F., Dong, H., Zhang, J., Wang, L., et~al. (2023).
\newblock Taskweaver: A code-first agent framework.
\newblock {\em arXiv preprint arXiv:2311.17541}.

\bibitem[Shen et~al., 2024]{shen2024hugginggpt}
Shen, Y., Song, K., Tan, X., Li, D., Lu, W., and Zhuang, Y. (2024).
\newblock Hugginggpt: Solving ai tasks with chatgpt and its friends in hugging face.
\newblock {\em Advances in Neural Information Processing Systems}, 36.

\bibitem[Steffensen et~al., 2016]{PSAAM}
Steffensen, J.~L., Dufault-Thompson, K., and Zhang, Y. (2016).
\newblock Psamm: A portable system for the analysis of metabolic models.
\newblock {\em PLOS Computational Biology}, 12(2):1--29.

\bibitem[Sun et~al., 2024]{sun2024lambda}
Sun, M., Han, R., Jiang, B., Qi, H., Sun, D., Yuan, Y., and Huang, J. (2024).
\newblock Lambda: A large model based data agent.
\newblock {\em arXiv preprint arXiv:2407.17535}.

\bibitem[Trirat et~al., 2024]{trirat2024automl}
Trirat, P., Jeong, W., and Hwang, S.~J. (2024).
\newblock Automl-agent: A multi-agent llm framework for full-pipeline automl.
\newblock {\em arXiv preprint arXiv:2410.02958}.

\bibitem[Tu et~al., 2023]{tu2023datascienceeducationlarge}
Tu, X., Zou, J., Su, W.~J., and Zhang, L. (2023).
\newblock What should data science education do with large language models?

\bibitem[Waller and Fawcett, 2016]{waller2016data}
Waller, M.~A. and Fawcett, S.~E. (2016).
\newblock Data science, predictive analytics, and big data: A revolution that will transform supply chain design and management.
\newblock {\em Journal of Business Logistics}, 37(1):55--62.

\bibitem[Wang et~al., 2024a]{wang2024data}
Wang, C., Lee, B., Drucker, S., Marshall, D., and Gao, J. (2024a).
\newblock Data formulator 2: Iteratively creating rich visualizations with ai.
\newblock {\em arXiv preprint arXiv:2408.16119}.

\bibitem[Wang et~al., 2024b]{wang2024enhancing}
Wang, J., Zhang, Z., He, Y., Zhang, Z., Song, Y., Shi, T., Li, Y., Xu, H., Wu, K., Yi, X., et~al. (2024b).
\newblock Enhancing code llms with reinforcement learning in code generation: A survey.
\newblock {\em arXiv preprint arXiv:2412.20367}.

\bibitem[Wang et~al., 2024c]{wang2024executable}
Wang, X., Chen, Y., Yuan, L., Zhang, Y., Li, Y., Peng, H., and Ji, H. (2024c).
\newblock Executable code actions elicit better llm agents.
\newblock {\em arXiv preprint arXiv:2402.01030}.

\bibitem[Wei et~al., 2022]{wei2022chain}
Wei, J., Wang, X., Schuurmans, D., Bosma, M., Xia, F., Chi, E., Le, Q.~V., Zhou, D., et~al. (2022).
\newblock Chain-of-thought prompting elicits reasoning in large language models.
\newblock {\em Advances in neural information processing systems}, 35:24824--24837.

\bibitem[Witten et~al., 2016]{witten2016data}
Witten, I.~H., Frank, E., and Hall, M.~A. (2016).
\newblock {\em Data Mining: Practical machine learning tools and techniques}.
\newblock Morgan Kaufmann.

\bibitem[Wu et~al., 2023]{wu2023autogen}
Wu, Q., Bansal, G., Zhang, J., Wu, Y., Zhang, S., Zhu, E., Li, B., Jiang, L., Zhang, X., and Wang, C. (2023).
\newblock Autogen: Enabling next-gen llm applications via multi-agent conversation framework.
\newblock {\em arXiv preprint arXiv:2308.08155}.

\bibitem[Xi et~al., 2023]{xi2023rise}
Xi, Z., Chen, W., Guo, X., He, W., Ding, Y., Hong, B., Zhang, M., Wang, J., Jin, S., Zhou, E., et~al. (2023).
\newblock The rise and potential of large language model based agents: A survey.
\newblock {\em arXiv preprint arXiv:2309.07864}.

\bibitem[Xie et~al., 2024]{xie2024waitgpt}
Xie, L., Zheng, C., Xia, H., Qu, H., and Zhu-Tian, C. (2024).
\newblock Waitgpt: Monitoring and steering conversational llm agent in data analysis with on-the-fly code visualization.
\newblock {\em arXiv preprint arXiv:2408.01703}.

\bibitem[Xie et~al., 2023]{xie2023openagents}
Xie, T., Zhou, F., Cheng, Z., Shi, P., Weng, L., Liu, Y., Hua, T.~J., Zhao, J., Liu, Q., Liu, C., et~al. (2023).
\newblock Openagents: An open platform for language agents in the wild.
\newblock {\em arXiv preprint arXiv:2310.10634}.

\bibitem[Yang et~al., 2024]{yang2024matplotagent}
Yang, Z., Zhou, Z., Wang, S., Cong, X., Han, X., Yan, Y., Liu, Z., Tan, Z., Liu, P., Yu, D., et~al. (2024).
\newblock Matplotagent: Method and evaluation for llm-based agentic scientific data visualization.
\newblock {\em arXiv preprint arXiv:2402.11453}.

\bibitem[Yao et~al., 2024]{yao2024tree}
Yao, S., Yu, D., Zhao, J., Shafran, I., Griffiths, T., Cao, Y., and Narasimhan, K. (2024).
\newblock Tree of thoughts: Deliberate problem solving with large language models.
\newblock {\em Advances in Neural Information Processing Systems}, 36.

\bibitem[Yao et~al., 2022]{yao2022react}
Yao, S., Zhao, J., Yu, D., Du, N., Shafran, I., Narasimhan, K., and Cao, Y. (2022).
\newblock React: Synergizing reasoning and acting in language models.
\newblock {\em arXiv preprint arXiv:2210.03629}.

\bibitem[Yi et~al., 2025]{yi2025score}
Yi, Q., He, Y., Wang, J., Song, X., Qian, S., Yuan, X., Sun, L., Xin, Y., Tang, J., Li, K., et~al. (2025).
\newblock Score: Story coherence and retrieval enhancement for ai narratives.
\newblock {\em arXiv preprint arXiv:2503.23512}.

\bibitem[Zhang et~al., 2024]{zhang2024ufo}
Zhang, C., Li, L., He, S., Zhang, X., Qiao, B., Qin, S., Ma, M., Kang, Y., Lin, Q., Rajmohan, S., et~al. (2024).
\newblock Ufo: A ui-focused agent for windows os interaction.
\newblock {\em arXiv preprint arXiv:2402.07939}.

\bibitem[Zhang et~al., 2023a]{zhang2023mlcopilot}
Zhang, L., Zhang, Y., Ren, K., Li, D., and Yang, Y. (2023a).
\newblock Mlcopilot: Unleashing the power of large language models in solving machine learning tasks.
\newblock {\em arXiv preprint arXiv:2304.14979}.

\bibitem[Zhang et~al., 2023b]{zhang2023data}
Zhang, W., Shen, Y., Lu, W., and Zhuang, Y. (2023b).
\newblock Data-copilot: Bridging billions of data and humans with autonomous workflow.
\newblock {\em arXiv preprint arXiv:2306.07209}.

\bibitem[Zhou et~al., 2024a]{zhou2024llm}
Zhou, X., Zhao, X., and Li, G. (2024a).
\newblock Llm-enhanced data management.
\newblock {\em arXiv preprint arXiv:2402.02643}.

\bibitem[Zhou et~al., 2024b]{zhou2024human}
Zhou, Z., Zhang, J., Zhang, J., He, Y., Wang, B., Shi, T., and Khamis, A. (2024b).
\newblock Human-centric reward optimization for reinforcement learning-based automated driving using large language models.
\newblock {\em arXiv preprint arXiv:2405.04135}.

\end{thebibliography}
}

\newpage
\begin{center}
    {\LARGE\bf Supplementary Materials for ``A Survey on Large Language Model-based Agents for Statistics and Data Science"}
\end{center}

\begin{abstract}
The supplementary materials provide details for the datasets used in main paper and additional case studies introduced in the main paper. Section \ref{sec:datasets} presents a summary table of all datasets used, including format, source, and analysis goals. Section \ref{sec:case_studies} describes additional case studies demonstrating the capabilities and limitations of LLM-based data agents in performing tasks such as visualization, regression diagnostics, and bootstrapped inference.
\end{abstract}

\noindent%
{\it Keywords: data agents; generative AI; data analysis; natural language interaction; statistical software.}  %3 to 6 keywords, that do not appear in the title
\vfill

\newpage
\spacingset{1.9} % DON'T change the spacing!

\section{Datasets}\label{sec:datasets}

\begin{table}[h]
\small
\centering
\begin{threeparttable}
\caption{Datasets used in the case studies, with format, availability, and analytical goals.}
\label{tab:datasets}
\begin{tabularx}{\textwidth}{@{}p{1.2cm} p{2.6cm} p{1.6cm} p{3.4cm} X@{}}
\toprule
\textbf{Case} & \textbf{Dataset} & \textbf{Format} & \textbf{Availability} & \textbf{Analysis Goal} \\
\midrule
1 & Wine Quality & CSV (4898, 11) & UCI Repository\tnote{1} & (1) Explore the effect of alcohol content on wine quality across red and white varieties. (2) Apply ML models and generate automatic reports. \\
\addlinespace
2 & Auto MPG & CSV (398, 7) & UCI Repository\tnote{5} & Perform linear regression to predict MPG and evaluate model assumptions using residual plots and the Breusch–Pagan test. \\
\addlinespace
3 & Wine Quality & CSV (4898, 11) & UCI Repository\tnote{1} & Estimate the mean alcohol content of red wine and construct a 95\% bootstrap confidence interval. \\
\addlinespace
4 & Salary Data & CSV (6750, 6) & Kaggle\tnote{3} & Visualize the average salary across different age groups using the Salary Data. \\
\addlinespace
5 & Breast Cancer Wisconsin & CSV (569, 30) & UCI Repository\tnote{4} & Train a binary classifier to predict malignant vs. benign tumors using diagnostic features. \\
\bottomrule
\end{tabularx}
\begin{tablenotes}
\footnotesize
\item[1] \url{https://archive.ics.uci.edu/ml/datasets/wine+quality}
\item[2] \url{https://archive.ics.uci.edu/dataset/109/wine}
\item[3] \url{https://www.kaggle.com/datasets/mohithsairamreddy/salary-data}
\item[4] \url{https://archive.ics.uci.edu/dataset/17/breast+cancer+wisconsin+diagnostic}
\item[5] \url{https://archive.ics.uci.edu/ml/datasets/auto+mpg}
\end{tablenotes}
\end{threeparttable}
\end{table}

\section{Additional Case Studies}\label{sec:case_studies}
\subsection{Case Study: Data Visualization and Machine Learning by Data Interpreter}

\begin{figure}
\centering
\includegraphics[width=\textwidth]{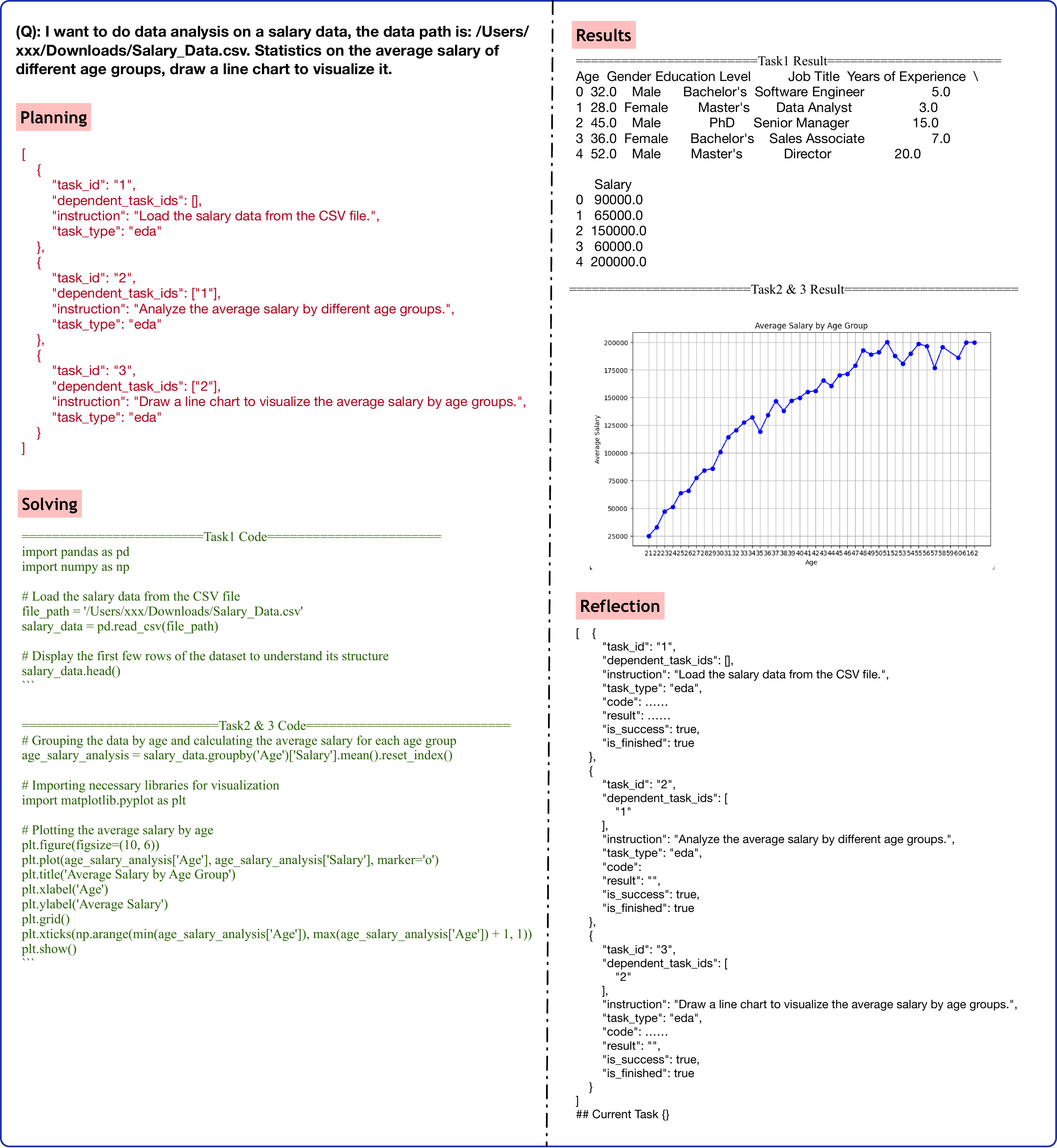}
\caption{Data Interpreter for the data visualization task.} \label{fig:di_vl}

\end{figure}

End-to-end data science agents are particularly convenient for users who wish to perform data-related tasks with a single prompt. In this case study, we demonstrate how an end-to-end data agent, the Data Interpreter, can handle both data visualization and machine learning tasks. Specifically, we prompt the Data Interpreter to visualize the average salary across different age groups using the Salary Data\footnote{\url{https://www.kaggle.com/datasets/mohithsairamreddy/salary-data}}. Figure \ref{fig:di_vl} shows both the prompt and the detailed internal process of the Data Interpreter in executing this task.

We pre-downloaded the data to disk due to errors that may occur during the download process. The Data Interpreter began by generating a path in the Planning phase, which consisted of three task nodes: (1) ``Load the salary data from the CSV file," (2) ``Analyze the average salary by different age groups," and (3) ``Draw a line chart to visualize the average salary by age groups." It then sequentially executed each task. After obtaining the results for each step, the interpreter checked whether the current task was successfully completed. If not, it re-executed the task. This reflection process ensures task completion and accuracy by allowing for iterative re-evaluation. Once all nodes are successfully completed, the overall task is considered finished.

\begin{figure}
\centering
\includegraphics[width=\textwidth]{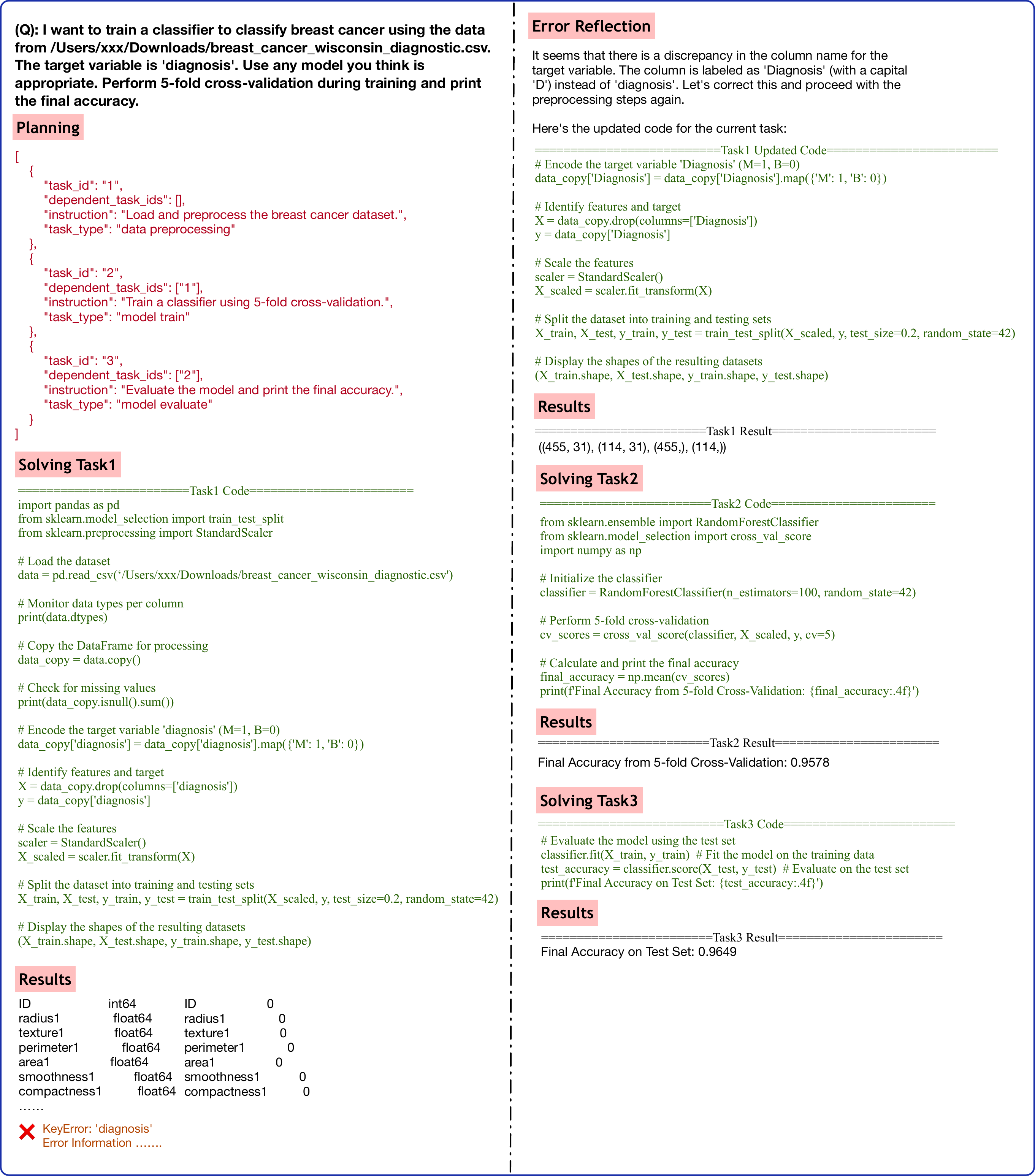}
\caption{Data Interpreter for the machine learning task.} \label{fig:di_ml}

\end{figure}

Then, we used the Data Interpreter to train a classifier for breast cancer prediction based on the Breast Cancer Wisconsin (Diagnostic) Dataset\footnote{\url{https://www.kaggle.com/datasets/uciml/breast-cancer-wisconsin-data}}. We prompted the Data Interpreter to compute the classification accuracy using 5-fold cross-validation. Figure \ref{fig:di_ml} illustrates the detailed prompt and problem-solving process.

Similarly, the Data Interpreter planned the whole task in 3 sub-tasks: (1) ``Load and preprocess the breast cancer dataset", (2) ``Train a classifier using 5-fold cross-validation.", and (3) ``Evaluate the model and print the final accuracy.". However, there was an error caused by the wrong column name when solving the first task. The Data Interpreter reflected the error and updated the task code. Eventually, it successfully finished all 3 sub-tasks and provided the final accuracy of 0.9649.

However, if the user's prompt is relatively simple, such as tasks that could be completed in a single step, the end-to-end data agent may waste many tokens due to its multi-step decomposition process.

Furthermore, because the entire process cannot be intervened by the user, if any step produces an undesired outcome, the entire workflow must be repeated during the next session. This will result in the waste of time and tokens. Additionally, users often overlook some details in their prompts. For example, in this case, we asked the agent to train a model but forgot to instruct it to save the model. As a result, the model was not saved, and the entire process needs to be repeated in the next session. Moreover, without setting a fixed \texttt{random\_state}, it is likely that we cannot reproduce the same model, leading to further inconsistencies.

To tackle this, the Data Interpreter offers another conversational mode for human interaction by specifying \texttt{--auto\_run False}. This hybrid approach, which supports both end-to-end and conversational modes, is likely to become a prevailing design trend in the future.

\subsection{Case Study: Residual Diagnostics and Heteroskedasticity Testing with GPT-4o}

\begin{figure}[h]
  \centering
  \includegraphics[width=0.9\textwidth]{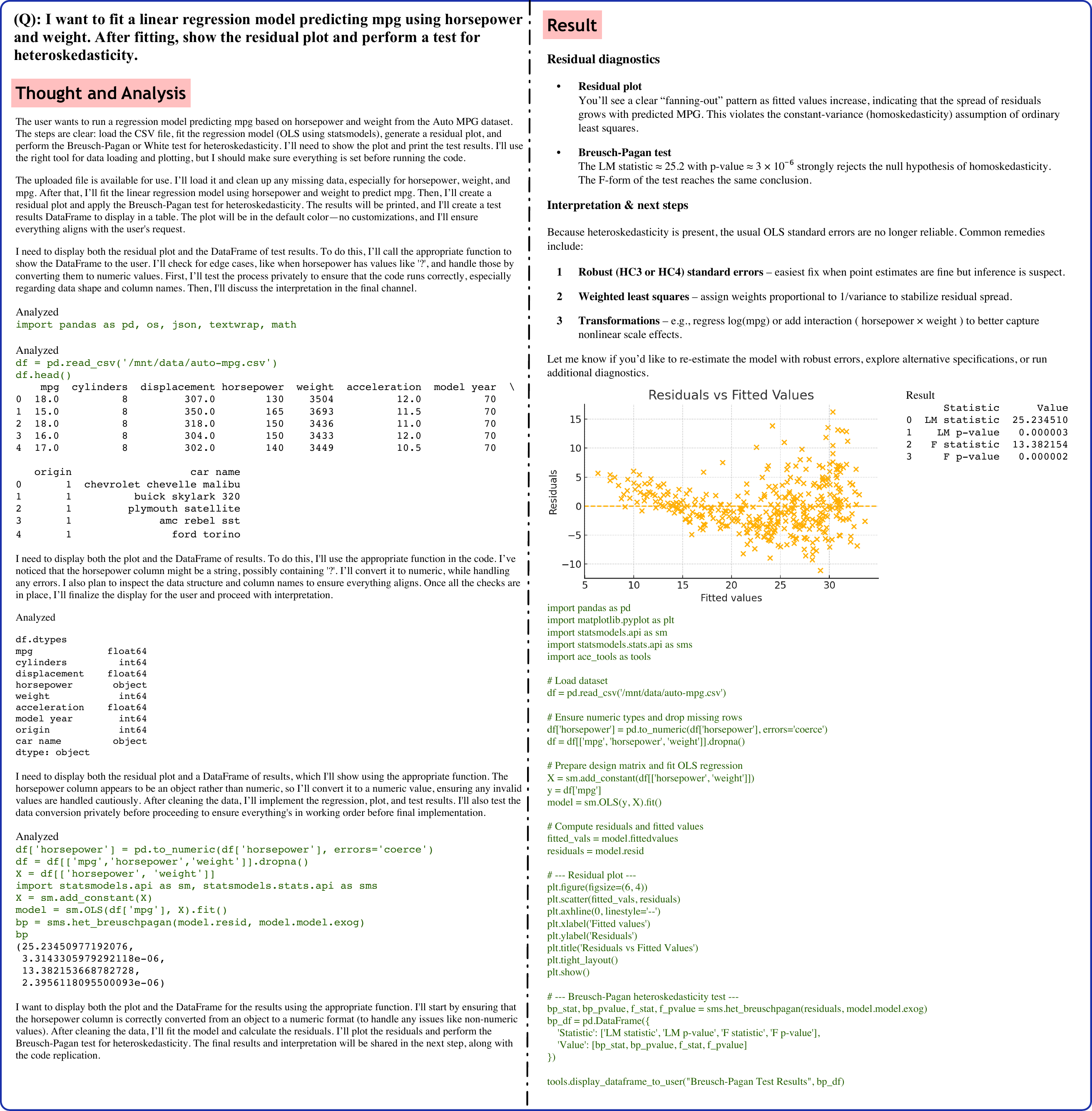}
  \caption{Dialogue of residual diagnostics and heteroskedasticity by GPT-4o.}
  \label{fig:residuals}
\end{figure}

To examine the ability of LLM-based data agents to perform statistically rigorous regression diagnostics, we prompted GPT-4o to conduct a linear regression analysis using the Auto MPG dataset.\footnote{Available from the UCI Machine Learning Repository: \url{https://archive.ics.uci.edu/ml/datasets/auto+mpg}} The goal was to predict \texttt{mpg} (miles per gallon) based on vehicle characteristics, notably \texttt{horsepower} and \texttt{weight}. The prompt and agent response are detailed in the figure \ref{fig:residuals}.

GPT-4o generated Python code using \texttt{statsmodels} to fit the model and perform residual diagnostics, including a Breusch–Pagan test for heteroskedasticity. The residual plot showed a clear “fanning-out” pattern, suggesting increasing variance in residuals with higher fitted values.

The Breusch–Pagan test returned an LM statistic of approximately 25.2 with a p-value near \(3 \times 10^{-6}\), strongly rejecting the null hypothesis of homoskedasticity. Besides, GPT-4o correctly noted that the presence of heteroskedasticity invalidates classical OLS inference and give recommendations.

This case demonstrates that LLM-based agents can not only execute regression modeling but also interpret key assumptions and suggest statistically sound next steps.

\subsection{Case Study: Bootstrap Confidence Interval Estimation with GPT-4o}

\begin{figure}[h]
  \centering
  \includegraphics[width=0.9\textwidth]{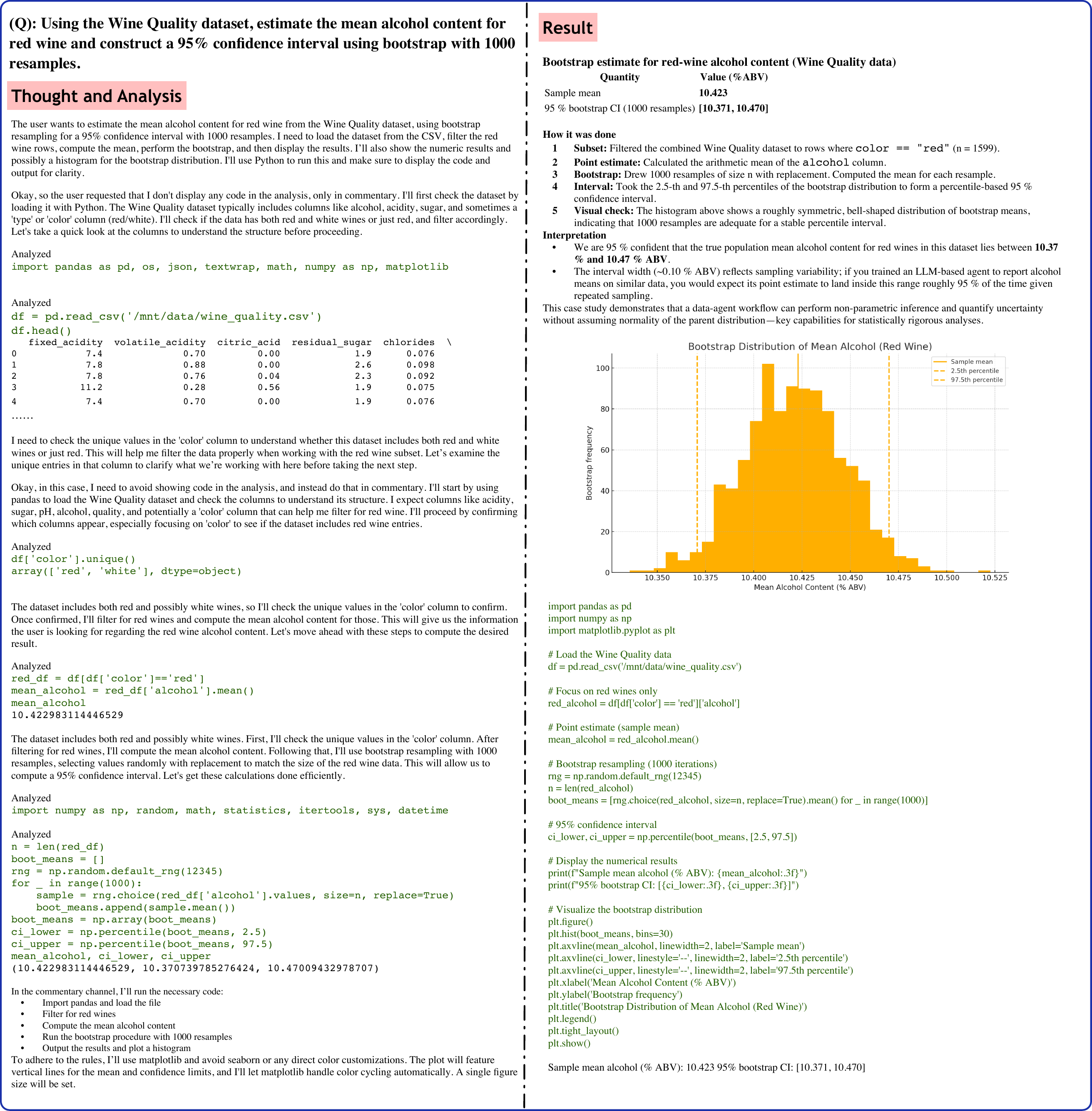}
  \caption{Response of GPT-4o for bootstrap confidence interval estimation.}
  \label{fig:cs_bootstrap}
\end{figure}

In this case study, we assessed whether GPT4-o can perform non-parametric inference through bootstrap resampling. We used the Wine Quality dataset, which contains physicochemical attributes of red and white wines. The task was to estimate the average alcohol content of red wine and construct a 95\% confidence interval using bootstrap. Figure \ref{fig:cs_bootstrap} shows the prompt and response from GPT-4o.

GPT-4o filtered the dataset to red wines (\(n = 1599\)) and computed the sample mean of the \texttt{alcohol} variable. It then performed 1000 bootstrap resamples, each of size 1599 drawn with replacement, and calculated the mean for each. The 2.5th and 97.5th percentiles of the resulting distribution were used to form the confidence interval.

GPT-4o noted that the bootstrap distribution was approximately symmetric and bell-shaped, suggesting adequate stability with 1000 resamples. The agent correctly interpreted that we are 95\% confident the true mean alcohol content for red wines lies within the estimated interval. The narrow width of the interval (\(\sim 0.1\%\) ABV) reflects relatively low sampling variability.

This case highlights that data agents can go beyond point estimation and engage in uncertainty quantification using robust, model-free methods such as the bootstrap. The approach avoids distributional assumptions, making it especially useful for inference in non-Gaussian settings.

However, for some commercial products, such as ChatGPT and ChatGLM, there may be restrictions when attempting to install additional packages. Due to the server's protection mechanisms, the sandbox environment does not have internet access, and the agent receives a notification stating, "There is no internet access." Any attempt to install packages results in an error. Details can be found in the supplementary materials. Figure \ref{fig:error_chatgpt} illustrates this issue.

\begin{figure}
        \centering
        \includegraphics[width=\textwidth]{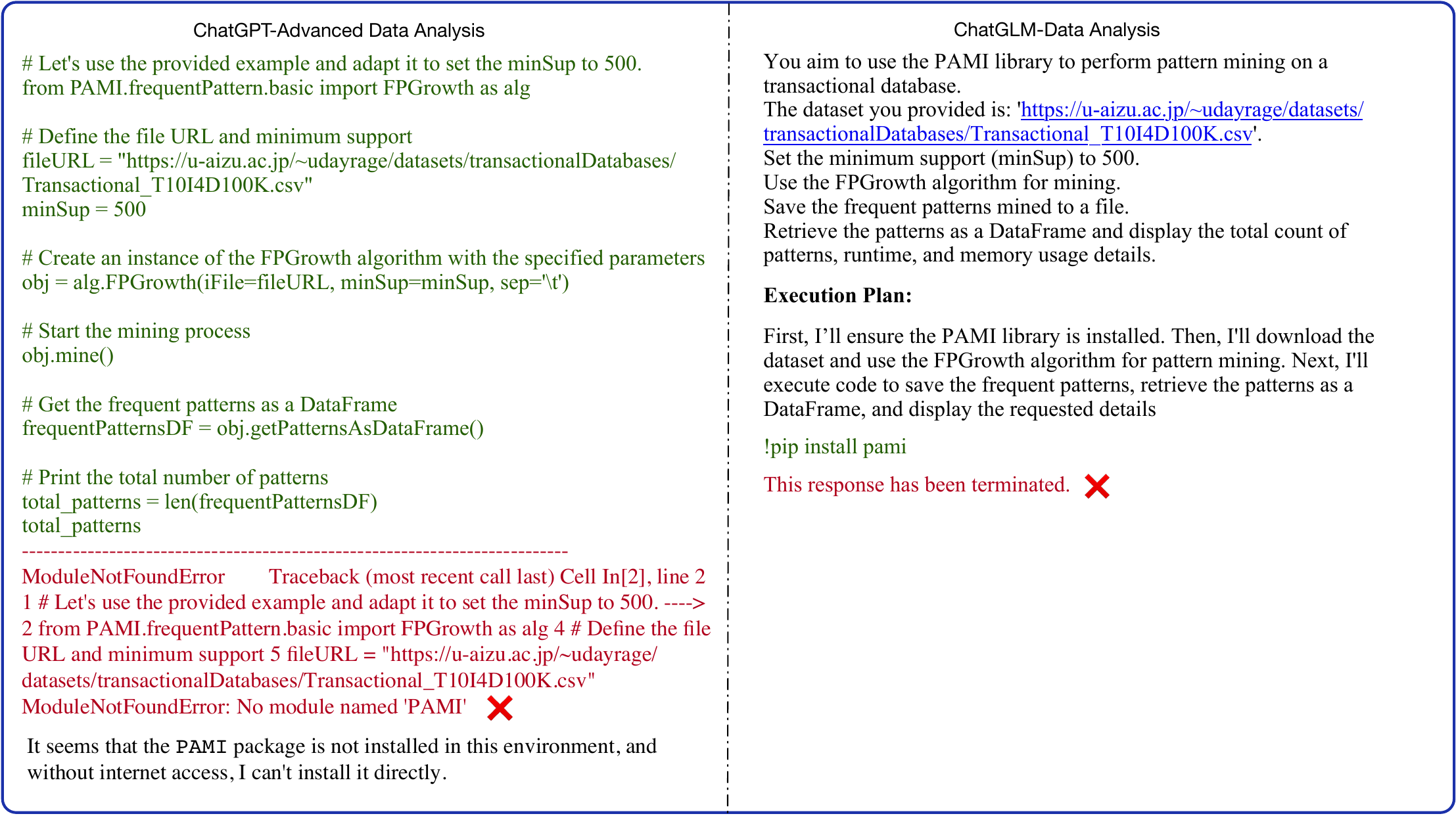}
        \caption{Scenarios when the task needs to install additional packages. The protection mechanism will prevent the installation and further terminate the task. Illustrated by ChatGPT-ADA and ChatGLM-DA.}
        \label{fig:error_chatgpt}
\end{figure}

% \newpage
% %\bigskip
% \begin{center}
% \textbf{\Large Supplementary Materials}
% \end{center}

\end{document}